\documentclass[twoside,11pt]{article}

\usepackage{jmlr2e}
\usepackage{amsmath}
\usepackage{graphicx}
\usepackage{caption}
\usepackage{subcaption}
\usepackage{epsfig} 
\usepackage{algorithmic}
\usepackage{algorithm}
\usepackage{tikz}
\usetikzlibrary{fit,positioning}
\usepackage{pgfplots}
\usepackage{hyperref}
\usepackage{url}
\usepgfplotslibrary{patchplots,colormaps}

\newcommand{\approptoinn}[2]{\mathrel{\vcenter{\offinterlineskip\halign{\hfil$##$\cr#1\propto\cr\noalign{\kern2pt}#1\sim\cr\noalign{\kern-2pt}}}}}
\newcommand{\appropto}{\mathpalette\approptoinn\relax}
\newcommand{\argmax}[1]{\underset{#1}{\operatorname{arg}\,\operatorname{max}}\;}

\ShortHeadings{A `Gibbs-Newton' Technique for Enhanced Inference of Topic Models}{Osama and David and Mike}
\firstpageno{1}

\begin{document}

\title{A `Gibbs-Newton' Technique for Enhanced Inference of Multivariate Polya Parameters and Topic Models}

\author{\name Osama Khalifa \email ok32@hw.ac.uk \\
       \name David Wolfe Corne \email d.w.corne@hw.ac.uk \\
       \name Mike Chantler \email m.j.chantler@hw.ac.uk \\
       \addr Mathematical and Computer Science \\
       Heriot-Watt University \\
       Riccarton, Edinburgh \\
       United Kingdom
}

\editor{}

\maketitle

\begin{abstract}
Hyper-parameters play a major role in the learning and inference process of latent Dirichlet allocation (LDA).
In order to begin the LDA latent variables learning process, these hyper-parameters values need to be pre-determined. 
We propose an extension for LDA that we call `Latent Dirichlet allocation Gibbs Newton' (LDA-GN), which places non-informative priors over these hyper-parameters and uses Gibbs sampling to learn appropriate values for them. 
At the heart of LDA-GN is our proposed `Gibbs-Newton' algorithm, which is a new technique for learning the parameters of multivariate Polya distributions. 
We report Gibbs-Newton performance results compared with two prominent existing approaches to the latter task: Minka's fixed-point iteration method and the Moments method. 
We then evaluate LDA-GN in two ways: (i) by comparing it with standard LDA in terms of the ability of the resulting topic models to generalize to unseen documents; (ii)  by comparing it with standard LDA in its performance on a binary classification task. 
\end{abstract}


\section{Introduction}
Large text corpora are increasingly abundant as a result of ever-speedier computational processing capabilities and ever-cheaper means of data storage. 
This has led to increased interest in the automated extraction of useful information from such corpora, and particularly in the task of automated characterization and/or summarization of each document in a corpus, as well as the corpus as a whole. 

It is generally tacitly understood that the first step in characterizing or describing an individual document is to identify the topics that are covered in that document.
Thus, there is much current research into topic modelling methods such as in \citep{Hofmann99,Blei03,Griffiths04,Buntine04}; these are algorithms that extract structured semantic topics from a collection of documents. These algorithms have many applications in various fields such as genetics \citep{Pritchard00}, image analysis \citep{FeiFei05,Russell06,Barnard03}, survey data processing \citep{Erosheva07} and social media analysis \citep{Airoldi07}.
Most current topic modelling methods are based on the well-known `bag-of-words' representation; in this approach, a document is represented simply as a bag of words, where words counts are preserved but their order in the original document is ignored.
Current topic modelling methods also tend to use probabilistic models, involving many observed and hidden variables which need to be learned from training data.

Latent Dirichlet allocation (LDA) \citep{Blei03}---the springboard for many other topic modelling methods---is the simplest topic modelling approach, and the most common one in use. However, pre-determined hyper-parameters play a major role in LDA's learning and inference process; most authors, whether they use LDA or other algorithms, use fixed hyper-parameter values.

In this paper, a new extension for LDA is proposed, which removes the need to pre-determine the hyper-parameters.
The basic idea behind this new version of LDA, which we call `Latent Dirichlet allocation Gibbs Newton' (LDA-GN), is to place non-informative uniform priors over the LDA hyper-parameters $\alpha$ and $\beta$. 
Each component in $\alpha$ and $\beta$ is sampled from a uniform distribution.  
A non-informative prior is used since we generally do not have prior information about these parameters.
We evaluate LDA-GN by comparing it with standard LDA using its recommended settings for $\alpha$ and $\beta$ as described and tested in \citep{Wallach09Rethinking}. 
This comparison is based on two evaluation metrics. 
Firstly, the perplexity of the inferred topic model, measured on unseen test documents (this is a common approach in the literature to evaluate topic models);
secondly, we test the inferred topic model's performance on a supervised task such as spam filtering.

At the heart of LDA-GN is what we call a 'Gibbs-Newton' (GN) approach for learning the hyper-parameters of a multivariate Polya distribution. 
Within LDA-GN, the role of GN is to learn the parameters for what amounts to the combination of two distinct multivariate Polya distributed data streams that are assumed in the LDA model. 
However, we also extract GN as a standalone method---since it is able to learn the parameters for any data distributed under a multivariate Polya distribution---and we compare it with two prominent methods for this task: the Moments method suggested by \citet{Ronning89} and Minka's fixed-point iteration method \citep{Minka00} enhanced by \citet{Wallach08}. A Java implementation for LDA-GN and also for standalone GN is provided at \href{http://is.gd/GNTMOD}{\url{http://is.gd/GNTMOD}}.

The rest of this paper is organised as follows: Firstly, for completeness, we provide a brief description of standard LDA; this includes an elaboration of the inference and learning algorithms involved in standard LDA, and some discussion of the effect of the hyper-parameters.  
Following that, we briefly review current algorithms for learning the parameters of multivariate Polya distributions. 
In the succeeding section, we present the proposed GN algorithm, and we evaluate it by comparison with other methods in terms of accuracy and speed.
Afterwards, the proposed new LDA extension, LDA-GN, is detailed. 
We then move on to a discussion of our evaluation metrics, followed by an evaluation of LDA-GN, before a concluding discussion.  

\section{An overview of Latent Dirichlet Allocation}
Latent Dirichlet Allocation (LDA) \citep{Blei03} is an unsupervised generative model to discover hidden topics in a collection of documents or corpus $W = \{W_1, W_2, .., W_M\}$ where $M$ is the total number of documents in the corpus $W$ and $W_i$ is the $i^{th}$ document in this corpus.
This model treats words as observed variables and topics as unobserved latent variables. 
The basic idea is to consider each document $W_d$ as having been generated by sampling from a mixture of latent topics $\theta_d$. 
Each topic $\varphi_k$ is a multinomial distribution over the full set of vocabulary terms in the corpus. 
By `vocabulary terms' we mean the list of unique word tokens that appear in the corpus.

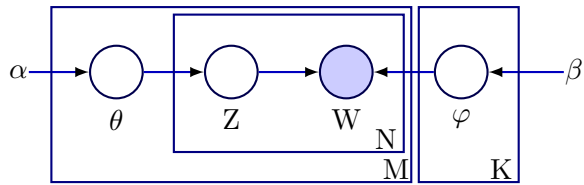
\begin{figure}
	\centering
	\begin{tikzpicture}
		\tikzstyle{variable}=[circle, minimum size = 7mm, thick, draw=blue!30!black, node distance = 8mm,label=below:#1]
		\tikzstyle{hyperparam}=[rectangle, label=center:#1]
		\tikzstyle{connect}=[-latex, thick, draw=blue!75!black]
		\tikzstyle{plate}=[rectangle, draw=blue!50!black, thick, label={[xshift=-15pt,yshift=13pt]south east:#1}]
		\node[hyperparam=$\alpha$] (alpha) { };
		\node[variable=$\theta$] (theta) [right=of alpha] { };
		\node[variable=Z] (Z) [right=of theta] {};
		\node[variable=W, fill = blue!20!white] (W) [right=of Z] { };
		\node[variable=$\varphi$] (phi) [right=of W] { };
		\node[hyperparam=$\beta$] (beta) [right=of phi] { };
		\path (alpha) edge [connect] (theta) (theta) edge [connect] (Z) (Z) edge [connect] (W) (phi) edge [connect] (W) (beta) edge [connect] (phi);
		\node[plate=N, inner xsep=4.0mm, inner ysep=5.5mm, fit = (Z) (W), yshift=-1.5mm] {};
		\node[plate=M, inner xsep=5.0mm, inner ysep=8.0mm, fit = (theta) (Z) (W), xshift=0mm, yshift=-3mm] {};
		\node[plate=K, inner xsep=3.0mm, inner ysep=8.0mm, fit = (phi), xshift=1mm, yshift=-3mm] {};
	\end{tikzpicture}
	\caption{LDA model}
	\label{fig:LDA}
\end{figure}

Dirichlet distributions $Dir(\theta | \alpha)$ and $Dir(\varphi | \beta)$ are used to model $\theta$ and $\varphi$ variables respectively. 
Where, $\alpha$ and $\beta$ are the concentration parameters for Dirichlet distributions and the model's hyper-parameters.
Figure \ref{fig:LDA} shows a graphical representation of the LDA model using plate notation where $K$ is the total number of latent topics and $Z$ is topic assignments for each word in each document.

Thus, the LDA model's generative process can be described as follows: at the beginning of the process, $K$ topic distributions $\varphi$ are sampled from a Dirichlet distribution over all vocabulary terms with parameter $\beta$. 
This set of topics is then used to represents all documents in the corpus.
For each document $W_d$ in the corpus, a topic mixture $\theta_d$ is sampled from a Dirichlet distribution with parameter $\alpha$. 
Then, in order to sample $W_{d,t}$ the $t^{th}$ word in the document $W_d$, a topic $Z_{d,t}$ is sampled from a multinomial distribution with parameter $\theta_d$. 
After that, the word $W_{d,t}$ itself is sampled from the multinomial distribution with parameter $\varphi_{Z_{d,t}}$ which is the word distribution of topic $Z_{d,t}$. 
 
Consequently, there are two Dirichlet distributions: 
The first one is a distribution over the $K$ dimensional simplex, which is used to model topic mixtures. 
On the other hand, the second one is a distribution over the $V$ dimensional simplex, which is used to model topic distributions; where $V$ is the total number of corpus vocabulary terms. 
LDA assumes that samples for each specific variable are considered to be i.i.d.

\subsection{LDA Learning and Inference}
From Figure \ref{fig:LDA} and the generative process described above, the LDA joint distribution is given by the following equation:
\begin{equation}
	P(W,Z,\theta,\varphi|\alpha,\beta)=  \prod_{k=1}^K {P(\varphi_k|\beta) \prod_{d=1}^M {P(\theta_d|\alpha) \prod_{t=1}^{N_d} P(Z_{d,t}|\theta_d)P(W_{d,t}|\varphi_{Z_{d,t}})}}\;\;, \nonumber
\end{equation}
\noindent
where $N_d$ is number of words in the document $W_d$. The conjugacy between Dirichlet and multinomial distributions allows $\theta$ and $\varphi$ to be marginalized out:
\begin{equation}
	P(W,Z|\alpha,\beta)=  \prod_{d=1}^{M}\frac{B(n_{d,\circ}+\alpha)}{B(\alpha)}\cdot \prod_{k=1}^{K}\frac{B(n_{\circ}^k+\beta)}{B(\beta)}\;\;. \nonumber
\end{equation}
\noindent
where, $n_{d,\circ}$ is a vector of length $K$, and each component value $n_{d,\circ}^k$ represents number of words in document $W_d$ assigned to the topic $k$.
On the other hand, $n_{\circ}^k$ is a vector of length $V$; each component value $n_{\circ,r}^k$ represents the number of instances of term $r$ in the whole corpus that are assigned to topic $k$. 
$B(\alpha)$ is the Dirichlet distribution's normalization constant, which is a multivariate version of the bivariate Beta function, and is the normalizing constant of Beta distribution. 
The Dirichlet distribution's normalization constant is given by the following formula:
\begin{equation}
	B(\alpha) = \frac{\prod_{k=1}^{K}\Gamma(\alpha_k)}{\Gamma(\sum_{k=1}^{K} \alpha_k)}\;\;. \nonumber
\end{equation}
The key inference problem that needs to be calculated is the posterior distribution given by the formula: 
\begin{equation}
	P(Z|W,\alpha,\beta)=  \frac{P(W,Z|\alpha,\beta)}{P(W|\alpha,\beta)} = \frac{\prod_{d=1}^{M}\prod_{t=1}^{N_d} P(W_{d,t},Z_{d,t}|\alpha,\beta)}{\prod_{d=1}^{M}\prod_{t=1}^{N_d} \sum_{k=1}^{K} P(Z_{d,t}=k, W_{d,t}|\alpha,\beta)}\;\;. \nonumber
\end{equation}
Unfortunately, the exact calculation of the posterior distribution is generally intractable due to the denominator. 
Its calculation involves summing over all possible settings of the topic assignment variable $Z$. 
This number has an exponential value given by $K^N$ where $N=\sum_{d=1}^{M}N_d$ is the total number of corpus words. 
However, there are several  approximation algorithms to sample from the posterior distribution which can be used for LDA such as: variational inference methods \citep{Blei03,Hoffman13}, expectation propagation \citep{Minka02}, and Gibbs sampling \citep{Steyvers07,Griffiths04,Pritchard00}. 
Variational methods and Markov-chain Monte Carlo methods such as Gibbs sampling are  widely used in the literature. 
Gibbs sampling, despite being the slowest, is widely considered to provide the most accurate results \citep{Namata09}. 
However, \citet{Asuncion09} show that with appropriate values of hyper-parameters $\alpha$ and $\beta$ these methods provide almost the same level of accuracy.  In this paper, Gibbs sampling is used for all experiments. 

\subsubsection{The LDA Gibbs Sampler}
Gibbs sampling is a Markov-chain Monte Carlo (MCMC) algorithm, which can be seen as a special case of the Metropolis--Hastings algorithm \citep{Metropolis1953}. 
It can be used to obtain an observation sequence from a high-dimensional multivariate probability distribution. 
Consequently, the sequence can be used to approximate a marginal distribution for one or a subset of the model's variables. 
In addition, it can be used to compute an integral over one of the hidden variables, and consequently compute its expected value. 
Thanks to the conjugacy between Multinomial and Dirichlet distributions, a `collapsed' Gibbs sampler \citep{Liu1994} can be implemented for LDA, where the $\theta$ and $\varphi$ variables can be analytically integrated out before carrying out the Gibbs sampling process. 
This allows us to sample directly from the distribution $P(Z|W,\alpha,\beta)$ instead of the distribution $P(Z|\theta,\varphi,W,\alpha,\beta)$. 
This in turns reduces the number of hidden variables in the LDA model, and makes inference and learning faster. 

In order to build a Gibbs sampler for a model with one multidimensional hidden variable $x$ and observed variable $D$, full conditionals $P(x_i|x_{\neg i},D)$ need to be calculated, 
where, $x_{\neg i}$ represents all other dimensions of variable $x$ excluding dimension $i$. 
Thus, the Gibbs sampling process involves repetition of two steps:
\begin{enumerate}
  \item Choose a dimension $i$ (order is not important)
  \item Sample $x_i$ from distribution $P(x_i|x_{\neg i},D)$.
\end{enumerate}

For LDA, it is required to get samples from the posterior distribution $P(Z|W,\alpha,\beta)$, thus full conditional distributions $P(Z_{(d,t)}|Z_{\neg (d,t)},W,\alpha,\beta)$ should be defined,  where, $Z_{\neg(d,t)}$ represents topic assignments values for all corpus words excluding the $t^{th}$ word in document $W_d$. 
Assuming that the $t^{th}$ word in document $W_d$ is a word instance of term $v$, $W_{(d,t)}=v$ then:
\begin{equation}
	\begin{split}
  	P(Z_{(d,t)}=k|Z_{\neg(d,t)},W,\alpha,\beta) &= \frac{P(Z_{(d,t)}=k,Z_{\neg(d,t)},W|\alpha,\beta)}{P(Z_{\neg(d,t)},W_{\neg(d,t)}|\alpha,\beta)P(W_{(d,t)}|\alpha,\beta)} \\
  	&\propto \frac{P(Z_{(d,t)}=k,Z_{d,\neg t},W_d|\alpha,\beta)}{P(Z_{d,\neg t},W_{d,\neg t}|\alpha,\beta)} \\
  	&\propto (n^{k, \neg (d,t)}_{d,\circ} + \alpha_k)\frac{n^{k,\neg (d,t)}_{\circ,v} + \beta_{v}}{\sum_{r=1}^{V} n^{k,\neg (d,t)}_{\circ,r} + \beta_r}\;\;.\nonumber
  	\end{split}
\end{equation}
where, $W_{d,\neg t}$ is words of document $W_d$ excluding its $t^{th}$ word and  $Z_{d,\neg t}$ is $W_{d,\neg t}$ word's topic assignments. Also,
$n^{k, \neg (d,t)}_{d,\circ}$ is the number of words in document $W_d$ assigned to topic $k$ excluding the document's $t^{th}$ word, whereas $n^{k,\neg (d,t)}_{\circ,v}$ is the number of word instances of term $v$ assigned to topic $k$ from all corpus documents excluding the $t^{th}$ word in document $W_d$. 
Finally, we need to construct values of $\theta$ and $\varphi$ which correspond to setting $Z$. By definition, those two variables are distributed Multinomially  with Dirichlet priors. 
Thus, they are distributed by Dirichlet-Multinomial distribution as follows:
\begin{equation}
  	P(\theta_d|Z_d,\alpha) \sim Dir(n_{d,\circ} + \alpha) \nonumber
\end{equation}
\begin{equation}
  	P(\varphi_k|Z,\beta) \sim Dir(n^{k}_{\circ} + \beta)\;\;. \nonumber
\end{equation}
\noindent
where, $n_{d, \circ}$ is a vector of topics observation counts in the document $W_d$ and $n^{k}_{\circ}$ is a vector of term observation counts for topic $k$. Therefore, and using the expectation of the Dirichlet distribution, $\theta$ and $\varphi$ corresponding to the setting $Z$ are given by:
\begin{equation}
  	\theta^{k}_{d} = \frac{n^{k}_{d,\circ} + \alpha_k}{\sum_{i=1}^{K} n^{i}_{d,\circ} + \alpha_i} 
  	\label{eq:LDA_Gibbs4}
\end{equation}
\begin{equation}
  	\varphi^{v}_{k} = \frac{n^{k}_{\circ,v} + \beta_v}{\sum_{r=1}^{V} n^{k}_{\circ,r} + \beta_r}\;\;. 
  	\label{eq:LDA_Gibbs5}
\end{equation}
Consequently, LDA's collapsed Gibbs sampling algorithm is given by Algorithm \ref{alg:LDACGS}, where, $n_{\circ,\circ}^{k} = \sum_{r=1}^{V} n_{\circ,r}^{k}$ and $\beta_{\circ} = \sum_{r=1}^{V} \beta_r$
\begin{algorithm}                      
\caption{LDA collapse Gibbs sampler}          
\label{alg:LDACGS}    
\begin{algorithmic}                    
    \REQUIRE $W$ words of the corpus, $\alpha$ and $\beta$ the model parameters.
    \ENSURE $Z$ topic assignments, $\theta$ topics mixtures, $\varphi$ topics distributions.
    \STATE Randomly initialize $Z$ with integers $\in [1..K]$
    \REPEAT
    \FOR {$d = 1$ \TO $M$}
		\FOR {$t = 1$ \TO $N_d$} 
			\STATE $v \leftarrow W_{d,t}$; $k \leftarrow Z_{d,t}$
			\STATE $n_{d,\circ}^{k} \leftarrow n_{d,\circ}^{k} - 1$; $n_{\circ,v}^{k} \leftarrow n_{\circ,v}^{k} - 1$; $n_{\circ,\circ}^{k} \leftarrow n_{\circ,\circ}^{k} - 1$;
			\STATE $k \sim (n^{k}_{d,\circ} + \alpha_k)\frac{n^{k}_{\circ,v} + \beta_{v}}{n^{k}_{\circ,\circ} + \beta_{\circ}}$
			\STATE $Z_{d,t} \leftarrow k$
			\STATE $n_{d,\circ}^{k} \leftarrow n_{d,\circ}^{k} + 1$; $n_{\circ,v}^{k} \leftarrow n_{\circ,v}^{k} + 1$; $n_{\circ,\circ}^{k} \leftarrow n_{\circ,\circ}^{k} + 1$;
		\ENDFOR
	\ENDFOR
    \UNTIL {convergence}
    \STATE Calculate $\theta$ using Equation \ref{eq:LDA_Gibbs4}
    \STATE Calculate $\varphi$ using Equation \ref{eq:LDA_Gibbs5}
    \RETURN $Z$,$\theta$,$\varphi$
\end{algorithmic}
\end{algorithm} 

\subsection{LDA Model Hyper-parameters}
The hyper-parameters $\alpha$ and $\beta$ play a large role in learning and building high-quality topic models \citep{Asuncion09,Wallach09Rethinking,Hutter14}. 
Typically, symmetric values of $\alpha$ and $\beta$ are used in the literature. 
Using symmetric $\alpha$ values means that all topics have the same chance to be assigned to a fixed number of documents. 
Symmetric $\beta$ values mean that all terms---frequent and infrequent ones---have the same chance to be assigned to fixed number of topics. 
However, according to \citet{Wallach09Rethinking}, using asymmetric $\alpha$ and symmetric $\beta$ tends to give the best performance results in terms of the inferred model's ability to generalise to unseen documents. The hyper-parameters 
$\alpha$ and $\beta$ generally have a smoothing effect over multinomial variables and they control the sparsity of $\theta$ and $\varphi$ respectively. 
The sparsity of $\theta$ is controlled by $\alpha$; hence smaller $\alpha$ values make the model prefer to describe each document using a smaller number of topics. The sparsity of
$\varphi$ is controlled by $\beta$; hence smaller $\beta$ values makes the model reluctant to assign corresponding terms to multiple topics. 
Consequently, similar words with similar small $\beta$ values tend to be assigned to the same subset of topics.

\section{Estimation of Multivariate Polya Distribution Parameters}
The Multivariate Polya distribution, also known as the Dirichlet-Multinomial distribution, is a compound distribution.
Sampling from the multivariate Polya distribution involves sampling a vector $\rho$ from a $K$ dimensional Dirichlet distribution with parameter $\alpha$ and then drawing a set of discrete samples from a categorical distribution with parameter $\rho$.  
This process corresponds to the 'Polya urn' which comprises sampling with replacement from an urn containing coloured balls. 
Every time a ball is sampled, its colour is observed and it is replaced into the urn; then an additional ball with the same colour is added to the urn. 

An inspection of the LDA model reveals that the model comprises two multivariate Polya distributions to model the data. 
The first distribution is used to model the distribution of the documents over topics given multinomial counts. 
These counts represent the numbers of words assigned to each topic for each document.  
The second distribution  models the distribution of the topics over vocabulary terms, given multinomial counts of the word instances assigned to different topics in the corpus as a whole.
Thus, accurate methods to learn multivariate Polya distribution parameters can enhance the quality of LDA topic modelling at the level of documents over topics, as well as at the level of topics over vocabulary terms. 

Consider a set of data-counts vectors $\mathcal{D} = \{\pi_1, \pi_2, .. \pi_N\}$ where $\pi_{j}^{i}$ is the number of times the outcome was $i$ in the $j^{th}$ sample. 
Assuming that these data are distributed according to a multivariate Polya distribution with parameter $\alpha$,
the basic idea behind learning this parameter from data is to maximize the likelihood:

\begin{equation}
  \begin{split}
     \mathcal{L}(\alpha|\mathcal{D})= \prod_{j=1}^{N} P(\pi_j|\alpha) &= \prod_{j=1}^{N} \int_{\rho_j} P(\pi_j|\rho_j)P(\rho_j|\alpha) \,d\rho_j\\
     			 &= \prod_{j=1}^{N} \frac{\Gamma(\alpha_{\circ})}{\Gamma(\pi_j^{\circ} + \alpha_{\circ})} \prod_{i=1}^{K} \frac{\Gamma(\pi_j^i+\alpha_i)}{\Gamma(\alpha_i)}\;\;. \nonumber 
  \end{split}
\end{equation}
where, $\alpha_{\circ} = \sum_{i=1}^{K} \alpha_i$ and $\pi_j^{\circ} = \sum_{i=1}^{K} \pi_j^i$. 
The research literature is replete with methods to estimate multivariate Polya parameters; however, there is no exact closed form solution available \citep{Ronning89,Wicker08}. 
One of the most accurate methods is Minka's fixed-point iteration method \citep{Minka00}.
On the other hand, one of the fastest methods is the Moments method \citep{Minka00,Ronning89,Leeds89}. Both will now be briefly described and reviewed.

\subsection{The Moments Method}
The Moments method is an approximate maximum likelihood technique useful as an initialization step for other methods. 
It provides a fast way to learn approximations to Dirichlet or multivariate Polya distribution parameters directly from data.
The Moments method uses known formulae for the first and second moments of the distribution's density function to calculate the parameter. 
The first moment (mean) of the multivariate Polya density function is given by the following formula:
\begin{equation}
  E[\pi^{i}] = \pi^{\circ}\frac{\alpha_i}{\alpha_{\circ}}\;\;. 
  \label{eq:Ronning01}
\end{equation}
It is easy to calculate the empirical mean value from data counts. 
Consequently, all that is required is to calculate the value $\alpha_{\circ}$ in order to figure out the value of parameter $\alpha$. 
This can be done using the second moment (variance) value. 
Variance of one dimension is enough to calculate $\alpha_{\circ}$ \citep{Dishon80}:
\begin{equation}
  var[\pi^{i}] = \frac{E[\pi^{i}](\pi^{\circ}-E[\pi^{i}])(\pi^{\circ}+\alpha_{\circ})}{\pi^{\circ}(1+\alpha_{\circ})}\;\;, \nonumber
\end{equation}
gives:
\begin{equation}
  \alpha_{\circ} = \frac{\pi^{\circ}\left(E[\pi^{i}]\left(\pi^{\circ}-E[\pi^{i}]\right)-var[\pi^{i}]\right)}{\pi^{\circ}\left(var[\pi^{i}]-E[\pi^{i}]\right) + {E[\pi^{i}]}^{2}}\;\;. \nonumber
\end{equation}
However, \citet{Ronning89} suggests that using the first $K-1$ dimensions gives more accurate results:
\begin{equation}
  \log \alpha_{\circ} = \frac{1}{K-1} \sum_{i=1}^{K-1} \log \left( \frac{\pi^{\circ}\left(E[\pi^{i}]\left(\pi^{\circ}-E[\pi^{i}]\right)-var[\pi^{i}]\right)}{\pi^{\circ}\left(var[\pi^{i}]-E[\pi^{i}]\right) + {E[\pi^{i}]}^{2}}\right)\;\;. 
  \label{eq:Ronning02}
\end{equation}

\subsection{Minka's Fixed-Point Iteration Method}
The Basic idea behind Minka's fixed-point iteration method for maximizing the likelihood is as follows:
Starting from an initial guess of the multivariate Polya distribution parameter $\alpha$, a simple lower bound on the likelihood, which is tight on $\alpha$. is constructed. 
The maximum value of this new lower bound is calculated in closed form and becomes a new estimate of $\alpha$ \citep{Minka00}.
This process is repeated until convergence.
Thus, the objective is to maximise the likelihood function for the multivariate Polya distribution:
\begin{equation}
   \mathcal{L}(\alpha|\mathcal{D}) = \prod_{j=1}^{N} \frac{\Gamma(\alpha_{\circ})}{\Gamma(\pi_j^{\circ} + \alpha_{\circ})} \prod_{i=1}^{K} \frac{\Gamma(\pi_j^i+\alpha_i)}{\Gamma(\alpha_i)}\;\;.
  \label{eq:Minka1}
\end{equation}
The following two lower bounds can be used to facilitate the calculation of the maximum likelihood:
\begin{equation}
  \frac{\Gamma(x)}{\Gamma(\eta+x)} \geq \frac{\Gamma(\hat{x})}{\Gamma(\eta + \hat{x})} e^{(\hat{x}-x)(\Psi(\eta+\hat{x}) - \Psi(\hat{x}))}
  \label{eq:Minka2}
\end{equation}
and,
\begin{equation}
  \frac{\Gamma(\eta+x)}{\Gamma(x)} \geq \frac{\Gamma(\hat{x}+\eta)}{\Gamma(\hat{x})}  \left( \frac{x}{\hat{x}}\right)^{\hat{x}\left[\Psi(\hat{x}+\eta)-\Psi(\hat{x})\right]}\;\;.
  \label{eq:Minka3}
\end{equation}
where $\eta \in \mathbb{Z}_{\ge 0}$ is a positive integer, $\hat{x} \in \mathbb{R}_{>0}$ and $x \in \mathbb{R}_{>0}$ are strictly positive real numbers. The
$\Psi$ function is the first derivative of the loggamma function, known as the digamma function \citep{Davis72}: $$\Psi(x) = \frac {\partial [\log\Gamma(x)]} {\partial x}=\frac{\Gamma'(x)}{\Gamma(x)}$$ 
Substituting Equation \ref{eq:Minka2} and Equation \ref{eq:Minka3} in Equation \ref{eq:Minka1} Leads to:
\begin{equation}
  \mathcal{L}(\alpha|\mathcal{D}) \geq \prod_{j=1}^{N} e^{-\alpha_{\circ}\left[ \Psi(\pi_j^{\circ}+\alpha_{\circ}^{\star}) - \Psi(\alpha_{\circ}^{\star}) \right]} \prod_{i=1}^{K} \alpha_{i}^{\alpha_{i}^{\star}\left[ \Psi(\pi_j^{i}+\alpha_{i}^{\star}) - \Psi(\alpha_{i}^{\star}) \right]} \cdot C\;\;.
  \label{eq:Minka4}
\end{equation}
where $\alpha_i^{\star}$, $\alpha_{\circ}^{\star}$ are two real values close to the original values of $\alpha_{i}$ and $\alpha_{\circ}$ respectively. 
The values used here are the previous guess of $\alpha_i$ and $\alpha_{\circ}$. 
And, $C$ is a constant that comprises all terms which do not involve $\alpha$.
Thus, taking the logarithm of both sides of Equation \ref{eq:Minka4} leads to:
\begin{equation}
  \log \mathcal{L}(\alpha|\mathcal{D}) \geq \mathcal{F}(\alpha) + C\;\;, \nonumber
\end{equation}
where, $\mathcal{F}$ is a function given by the following formula:
\begin{equation}
  \mathcal{F}(\alpha) = \sum_{j=1}^{N} -\alpha_{\circ} \left[ \Psi(\pi_{j}^{\circ}+\alpha_{\circ}^{\star}) - \Psi(\alpha_{\circ}^{\star})\right] + \sum_{i=1}^{K} \log \alpha_i \left[ \Psi(\pi_j^i + \alpha_i^{\star}) - \Psi(\alpha_i^{\star})\right] \alpha_i^{\star}\;\;. \nonumber
\end{equation}
Consequently, it is possible find the maximum of this bound in a closed form. Firstly by calculating the derivative of $\mathcal{F}$ with respect to $\alpha_i$ and then solving the equation $ \frac{\partial [\mathcal{F}(\alpha)]}{\partial \alpha_i} = 0$:
\begin{equation}
  \begin{split}
     \frac {\partial[\mathcal{F}(\alpha)]}{\partial \alpha_i} &= \sum_{j=1}^{N} \frac{\left[ \Psi(\pi_j^i + \alpha_i^{\star}) - \Psi(\alpha_i^{\star})\right] \alpha_i^{\star}}{\alpha_i} - \left[ \Psi(\pi_{j}^{\circ}+\alpha_{\circ}^{\star}) - \Psi(\alpha_{\circ}^{\star})\right] \\ &= 0\;\;. \nonumber
  \end{split}
\end{equation}
The previous first degree equation has a simple solution:
\begin{equation}
  \alpha_i  = \alpha_i^{\star} \frac{\sum_{j=1}^{N} \Psi(\pi_j^i + \alpha_i^{\star}) - \Psi(\alpha_i^{\star})}{\sum_{j=1}^{N} \Psi(\pi_{j}^{\circ}+\alpha_{\circ}^{\star}) - \Psi(\alpha_{\circ}^{\star})}\;\;.
  \label{eq:Minka8}
\end{equation}

\citet{Wallach08} provides a faster version of this algorithm by using the digamma function recurrence relation.
This is done by representing data counts samples as histograms. 
In other words, Let $N$ be the number of samples for a $K$ dimensional multivariate Polya distribution. 
Then, a more efficient representation would be as $K$ vectors of counts of elements,  where the $m^{th}$ cell of the $i^{th}$ vector represents the number of times the count $m$ is observed in the set of $N$ values related to the dimension $i$.
This value is represented by: 
\begin{equation}
  C_i^m = \sum_{j=1}^{N} \delta(\pi_j^i-m)\;\;.
  \label{eq:Minka9}
\end{equation}
Similarly, $C_{\circ}^m$ represents the number of times the sum $m$ is observed in the set of $N$ sum values---over all dimensions---of sample counts. 
\begin{equation}
  C_{\circ}^m = \sum_{j=1}^{N} \delta \left(\pi_j^{\circ}-m \right)\;\;,
  \label{eq:Minka10}
\end{equation}
where, $\pi_j^{\circ} = \sum_{i=1}^{K} \pi_j^i$. 
Equation \ref{eq:Minka9}, Equation \ref{eq:Minka10} allows us to rewrite Equation \ref{eq:Minka8} in a more efficient way:
\begin{equation}
  \alpha_i  = \alpha_i^{\star} \frac{\sum_{m=1}^{dim(C_i)} C_i^m \left[\Psi(m + \alpha_i^{\star}) - \Psi(\alpha_i^{\star})\right]}{\sum_{m=1}^{dim(C_{\circ})} C_{\circ}^m \left[\Psi(m+\alpha_{\circ}^{\star}) - \Psi(\alpha_{\circ}^{\star})\right]}\;\;,
  \label{eq:Minka11}
\end{equation}
where, $C_i$ is a histogram vector for all counts in the $N$ samples associated to the dimension $i$ and $C_{\circ}$ is a histogram vector for all counts sums over all dimensions in the $N$ samples.
And, $dim(C_i)$ and $dim(C_{\circ})$ are the numbers of elements in vectors $C_i$ and $C_{\circ}$ respectively.
This new formula speeds the computation to an extent that depends on how many frequent count values can be spotted in each dimension $i \in [1..K]$. 
The more frequent values there are the faster the computation is. 
Unfortunately, the digamma function call is time-consuming in practice; however, \citet{Wallach08} suggests that there is  room for improving the performance by getting rid of the digamma function call completely. 
This can be done by taking into consideration the digamma recurrence relation in \citep{Davis72}:
\begin{equation}
  \Psi(x + 1) = \Psi(x) + \frac{1}{x}\;\;. \nonumber
\end{equation}
This formula can be extended for any positive integer $m$:
\begin{equation}
  \Psi(x + m) = \Psi(x) + \sum_{l=1}^{m} \frac{1}{x + l - 1}\;\;. \nonumber
\end{equation}
Rewriting gives:
\begin{equation}
  \Psi(x + m) - \Psi(x) = \sum_{l=1}^{m} \frac{1}{x + l - 1}\;\;.
  \label{eq:Minka14}
\end{equation}
Substituting Equation \ref{eq:Minka14} in Equation \ref{eq:Minka11} leads to:
\begin{equation}
  \alpha_i  = \alpha_i^{\star} \frac{\sum_{m=1}^{dim(C_i)} C_i^m \sum_{l=1}^{m} \frac{1}{\alpha_{i}^{\star} + l - 1}}{\sum_{m=1}^{dim(C_{\circ})} C_{\circ}^m \sum_{l=1}^{m} \frac{1}{\alpha_{\circ}^{\star} + l - 1}}\;\;. 
  \label{eq:Minka15}
\end{equation}
An efficient implementation of Minka's fixed-point iteration using Equation \ref{eq:Minka15} is described in Algorithm \ref{alg:Minka1}.

\begin{algorithm}                      
\caption{Minka fixed-point iteration method}          
\label{alg:Minka1}    
\begin{algorithmic}                    
    \REQUIRE $C$ samples counts histograms, $C_{\circ}$ samples lengths histogram.
    \ENSURE $\alpha$ the parameter for multivariate polya distribution.
    \STATE Initialize $\alpha$ using Equation \ref{eq:Ronning01} and Equation \ref{eq:Ronning02} (the Moments method).
    \REPEAT
		\STATE $Dgma \leftarrow 0$
		\STATE $Dntr \leftarrow 0$
		\FOR {$m = 1$ \TO $dim(C_{\circ})$}
			\STATE $Dgma  \leftarrow Dgma  + \frac{1}{\alpha_{\circ} + m - 1} $
			\STATE $Dntr \leftarrow Dntr + C_{\circ}^{m} Dgma$
		\ENDFOR
		\FOR {$i = 1$ \TO $K$} 
			\STATE $Dgma \leftarrow 0$
			\STATE $Nmtr \leftarrow 0$
			\FOR {$m = 1$ \TO $dim(C_{i})$}
				\STATE $Dgma \leftarrow Dgma + \frac{1}{\alpha_i + m - 1}$
				\STATE $Nmtr \leftarrow Nmtr + C_{i}^{m} Dgma$
			\ENDFOR
			\STATE $\alpha_i \leftarrow \alpha_i \frac{Nmtr}{Dntr}$
		\ENDFOR
    \UNTIL {convergence}
    \RETURN $\alpha$
\end{algorithmic}
\end{algorithm} 

\subsection{The Proposed GN Method}
\begin{figure}
\centering
	\begin{tikzpicture}
		\tikzstyle{variable}=[circle, minimum size = 7mm, thick, draw=blue!30!black, node distance = 8mm,label=below:#1]
		\tikzstyle{hyperparam}=[rectangle, label=center:#1]
		\tikzstyle{connect}=[-latex, thick, draw=blue!75!black]
		\tikzstyle{plate}=[rectangle, draw=blue!50!black, thick, label={[xshift=-14pt,yshift=13pt]south east:#1}]
		\node[hyperparam=$a$] (a) { };
		\node[variable=$\alpha$] (alpha) [right=of a] { };
		\node[variable=$\rho$] (rho) [right=of alpha,] { };
		\node[variable=$\pi$, fill = blue!20!white] (pi) [right=of rho] { };
		\path (a) edge [connect] (alpha) (alpha) edge [connect] (rho) (rho) edge [connect] (pi);
		\node[plate=K, inner xsep=3.0mm, inner ysep=8.0mm, fit = (alpha), xshift=-1mm, yshift=-3mm] {};
		\node[plate=N, inner xsep=4.0mm, inner ysep=8.0mm, fit = (rho) (pi), yshift=-3mm] {};
	\end{tikzpicture}
\caption{Polya distribution generative model}
\label{fig:Polya}
\end{figure}
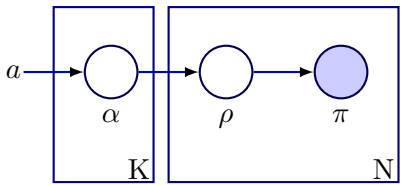

The parameters of a multivariate Polya distribution or Dirichlet distribution can be learned from data using standard Bayesian methods. In this paper, we focus on the multivariate Polya distribution as it plays a major role in LDA.
Thus, given $N$ samples from a multivariate Polya distribution, the data can be modelled using the generative model shown in Figure \ref{fig:Polya}.
The generative process in this case amounts to first sampling a value for each of the $K$ components from a uniform distribution with parameters 0,$a$. Then, a vector  $\rho$ of dimension $K$ is sampled from a Dirichlet distribution with parameter $\alpha$. 
Eventually, a multinomial variable $\pi$ is sampled from the multinomial distribution with parameter $\rho$.
A non-informative uniform prior is placed before each component $\alpha_i$ of the parameter vector $\alpha$ because we have no prior knowledge about their values.
The model's joint probability is:
\begin{equation}
	P(\alpha, \pi | a) = \prod_{j=1}^{N} \int_{\rho} P(\pi|\rho)P(\rho|\alpha) d\rho \prod_{i=1}^{K} P(\alpha | a)\;\;. \nonumber
\end{equation}
where $P(\pi|\rho) \sim Mutlnomial(\rho)$, $P(\rho|\alpha) \sim Dir(\alpha)$ and $P(\alpha | a) \sim Uniform(0,a)$. 
Probability densities substitution and further simplification leads to:
\begin{equation}
	P(\alpha, \pi | a) = \frac{1}{a^K}\prod_{j=1}^{N} \frac{\Gamma(\alpha_{\circ})}{\Gamma(\pi_j^{\circ} + \alpha_{\circ})} \prod_{i=1}^{K} \frac{\Gamma(\pi_j^i + \alpha_i)}{\Gamma(\alpha_i)}\;\;, \nonumber
\end{equation}
where, $\pi_j^i$ represents the count in sample $j$ and dimension $i$; whereas, $\pi_j^{\circ}$ represents count sum in sample $j$ over all dimensions.
In order to learn values of the hidden variable $\alpha$ a Gibbs sampler needs to be designed. 
The goal of Gibbs sampling here is to approximate the distribution $P(\alpha | \pi, a)$.
This takes place after calculating the distribution $P(\alpha_{k} | \alpha_{\neg k}, \pi, a)$ and then we sample each $\alpha_i$ value separately.
\begin{equation}
	\begin{split}
		P(\alpha_{k} | \alpha_{\neg k}, \pi, a) &= \frac{P(\alpha_{k}, \alpha_{\neg k}, \pi | a)}{P(\alpha_{\neg k}, \pi | a)} \\
		&\propto P(\alpha, \pi | a)\;\;. \nonumber
	\end{split}	
\end{equation}
It is not important to calculate the exact probability for Gibbs sampling. A ratio of probabilities is sufficient; thus, starting from the joint distribution:
\begin{equation}
	\begin{split}
		P(\alpha_{k} | \alpha_{\neg k}, \pi, a) &\propto \prod_{j=1}^{N} \frac{\Gamma(\alpha_{\circ})}{\Gamma(\pi_j^{\circ} + \alpha_{\circ})} \frac{\Gamma(\pi_j^k + \alpha_k)}{a\Gamma(\alpha_k)} \prod_{i \neq k} \frac{\Gamma(\pi_j^i + \alpha_i)}{\Gamma(\alpha_i)} \\
		&\propto \prod_{j=1}^{N} \frac{\Gamma(\alpha_{\circ})}{\Gamma(\pi_j^{\circ} + \alpha_{\circ})} \frac{\Gamma(\pi_j^k + \alpha_k)}{\Gamma(\alpha_k)}\;\;. \nonumber
	\end{split}
\end{equation}
Instead of sampling from this distribution, the value which maximize the logarithm of this density function is taken. 
Thus, the task is to maximize the function $\mathcal{F}(x)$ which is given by the following formula:
\begin{equation}
	\mathcal{F}(\alpha_k) =  \sum_{j=1}^{N} [\log \Gamma(\pi_j^k + \alpha_k) - \log \Gamma(\alpha_k)] - [ \log \Gamma(\pi_j^{\circ} + \alpha_{\circ}) - \log \Gamma(\alpha_{\circ})]\;\;. \nonumber
\end{equation}
The first derivative of $\mathcal{F}(\alpha_k)$ is:
\begin{equation}
	\begin{split}
		\frac{\partial[\mathcal{F}(\alpha_k)]}{\partial \alpha_k} &= \sum_{j=1}^{N} [\Psi(\pi_j^k + \alpha_k) - \Psi(\alpha_k)] - [\Psi(\pi_j^{\circ} + \alpha_{\circ}) - \Psi(\alpha_{\circ})] \\
		&= 0\;\;. 
	\end{split}
	\label{eq:GibbsNewton03}
\end{equation}
Unfortunately, there is no trivial solution for the previous equation; so we use Newton's method to find its root.
Although the previous equation has multiple roots, we are interested in the positive real root.
In order to apply Newton's method, we need to find the second derivative of $\mathcal{F}(\alpha_k)$.
\begin{equation}
	\frac{\partial^2 [\mathcal{F}(\alpha_k)]}{\partial \alpha_k^2} = \sum_{j=1}^{N} [ \Psi_1(\pi_j^k + \alpha_k) - \Psi_1(\alpha_k)] - [\Psi_1(\pi_j^{\circ} + \alpha_{\circ}) - \Psi_1\left(\alpha_{\circ}\right)]\;\;. \nonumber
\end{equation} 
where $$\Psi_1(x) = \frac{\partial^2 \log \Gamma(x)}{\partial x^2}$$ is the second derivative of the loggamma function, called the trigamma function \citep{Davis72}. 
It is not important to find a solution with high precision at the beginning, because it can be seen that Equation \ref{eq:GibbsNewton03} includes the coefficient $\alpha_{\circ} = \sum_{k=1}^{K} \alpha_k$. 
This coefficient is not accurate in the first iteration of Gibbs sampling as it represents a sum of guessed values.
The value of $\alpha_{\circ}$ is updated after each full iteration of the Gibbs sampler; in other words, after processing all $\alpha_k$ values. 
Thus, only one iteration of Newton's method is used for each $\alpha_{k}$.
\begin{equation}
	\alpha_k = \alpha_k^{\star} - \frac{\sum_{j=1}^{N} \left[\Psi(\pi_j^k + \alpha_k^{\star}) - \Psi(\alpha_k^{\star})\right] - \left[\Psi(\pi_j^{\circ} + \alpha_{\circ}^{\star}) - \Psi(\alpha_{\circ}^{\star})\right]}{\sum_{j=1}^{N} \left[ \Psi_1(\pi_j^k + \alpha_k^{\star}) - \Psi_1(\alpha_k^{\star})\right] - \left[\Psi_1(\pi_j^{\circ} + \alpha_{\circ}^{\star}) - \Psi_1(\alpha_{\circ}^{\star})\right]}\;\;. \nonumber
\end{equation} 
Rewriting by taking into consideration the recurrence formulae for the digamma and trigamma functions:
\begin{equation}
	\Psi(x+\eta)-\Psi(x) = \sum_{l=1}^{\eta} \frac{1}{(x + l - 1)} \nonumber
\end{equation} 
\begin{equation}
	\Psi_1(x+\eta)-\Psi_1(x) = \sum_{l=1}^{\eta} \frac{-1}{(x + l - 1)^2} \nonumber
\end{equation} 
gives:
\begin{equation}
	\alpha_k = \alpha_k^{\star} - \frac{\sum_{j=1}^{N} \sum_{l=1}^{\pi_j^k} \frac{1}{(\alpha_k^{\star} + l - 1)} -  \sum_{l=1}^{\pi_j^{\circ}}  \frac{1}{(\alpha_{\circ}^{\star} + l - 1)}}{\sum_{j=1}^{N} \sum_{l=1}^{\pi_j^k} \frac{-1}{(\alpha_k^{\star} + l - 1)^2} -  \sum_{l=1}^{\pi_j^{\circ}}  \frac{-1}{(\alpha_{\circ}^{\star} + l - 1)^2}}\;\;. \nonumber
\end{equation} 
Rewriting using the histogram counts $C$ for more computational efficiency:
\begin{equation}
	\alpha_k = \alpha_k^{\star} - \frac{L_1 - \sum_{m=1}^{dim(C_{k})} C_{k}^{m} \sum_{l=1}^{m} \frac{1}{(\alpha_k^{\star} + l - 1) \ }}{L_2 - \sum_{m=1}^{dim(C_{k})} C_{k}^{m} \sum_{l=1}^{m} \frac{-1}{(\alpha_k^{\star} + l - 1)^2}}\;\;. \nonumber
\end{equation} 
where $L_1$ and $L_2$ are given by the following formulae:
\begin{equation}
	L_1 = \sum_{m=1}^{dim(C_{\circ})} C_{\circ}^{m} \sum_{l=1}^{m}  \frac{1}{(\alpha_{\circ}^{\star} + l - 1)} \nonumber
\end{equation}
\begin{equation}
	L_2 = \sum_{m=1}^{dim(C_{\circ})} C_{\circ}^{m} \sum_{l=1}^{m}  \frac{-1}{(\alpha_{\circ}^{\star} + l - 1)^2}\;\;. \nonumber
\end{equation}
The complete GN method is described in Algorithm \ref{alg:GibbsNewton1}.
\begin{algorithm}                      
\caption{GN method pseudo code}          
\label{alg:GibbsNewton1}    
\begin{algorithmic}                    
    \REQUIRE $C$ samples counts histograms, $C_{\circ}$ samples lengths histogram.
    \ENSURE $\alpha$ the parameter for multivariate polya distribution.
    \STATE Initialize $\alpha$ using Equation \ref{eq:Ronning01} and Equation \ref{eq:Ronning02} (the Moments method).
    \REPEAT
		\STATE $Dgma \leftarrow 0$, $Tgma \leftarrow 0$
		\STATE $L_1 \leftarrow 0$, $L_2 \leftarrow 0$
		\STATE $\alpha_{\circ} \leftarrow \sum_{i=1}^K \alpha_i$
		\FOR {$m = 1$ \TO $dim(C_{\circ})$}
			\STATE $Dgma  \leftarrow Dgma  + \frac{1}{(\alpha_{\circ} + m - 1)} $
			\STATE $Tgma  \leftarrow Tgma  - \frac{1}{(\alpha_{\circ} + m - 1)^2} $
			\STATE $L_1 \leftarrow L_1 + C_{\circ}^{m} Dgma$
			\STATE $L_2 \leftarrow L_2 + C_{\circ}^{m} Tgma$			
		\ENDFOR
		\FOR {$i = 1$ \TO $K$} 
			\STATE $Dgma \leftarrow 0$, $Tgma \leftarrow 0$
			\STATE $Nmtr \leftarrow 0$, $Dntr \leftarrow 0$
			\FOR {$m = 1$ \TO $dim(C_{i})$}
				\STATE $Dgma \leftarrow Dgma + \frac{1}{(\alpha_i + m - 1)}$
				\STATE $Tgma \leftarrow Tgma - \frac{1}{(\alpha_i + m - 1)^2}$
				\STATE $Nmtr \leftarrow Nmtr + C_{i}^{m} Dgma$
				\STATE $Dntr \leftarrow Dntr + C_{i}^{m} Tgma$
			\ENDFOR
			\STATE $\alpha_i^{new} \leftarrow \alpha_i - \frac{L_1 - Nmtr}{L_2 - Dntr}$
			\IF {$\alpha_i^{new} < 0$}
				\STATE $\alpha_i^{new} \leftarrow \frac{\alpha_i}{2} $ 
			\ENDIF
			\STATE $\alpha_i \leftarrow \alpha_i^{new}$
		\ENDFOR
    \UNTIL {convergence}
    \RETURN $\alpha$
\end{algorithmic}
\end{algorithm} 

\subsection{Experimental Evaluation: GN}
In this section, two main experiments are designed to assess the performance of the GN method against the Moments method and Minka's fixed-point iteration. 
The first experiment is intended to evaluate accuracy whereas the second experiment is aimed at assessing its efficiency.
Artificial data are used in this section, allowing us to compare methods under a wide variety of conditions.  
The number of multivariate Polya samples used ranges from 10 to 1000, and the number of elements used to generate each sample falls in the range $\left[1000, 20000\right]$.

\subsubsection{Accuracy Benchmark}
In order to assess the accuracy of the proposed method, two categories of data sets are considered. 
Both categories are designed to use a 10 dimensional multivariate Polya distribution with known parameters $\alpha$. 
The first category has small component values in $\alpha$, being  real numbers sampled uniformly from the range $\left]0, 1\right]$.
The second category has relatively large $\alpha$ component values, in the range $\left]0,50\right]$, again sampled uniformly.
Each category can contain from 50 to 1000 multinomial count vectors or multivariate Polya samples.
 
The Moments method, Minka's fixed-point iteration method, and the proposed GN method are used to learn the parameter $\alpha$ vectors from the data. Given the resulting $\alpha$ vector, the difference between each component $\alpha_i$ and its actual value is calculated and registered. 
80 experiments were done, 40 for each category of data set, allowing these methods to be evaluated under highly varied settings in terms of data sparsity and number of samples needed.
Figure \ref{fig:accuracy10Dim_1} displays the differences of small $\alpha$ components and their actual values using the first set of data. 
Figure \ref{fig:accuracy10Dim_50} shows the differences when $\alpha_i$  has relatively large values, in the second category of data.
The figure indicates that Minka's fixed-point iteration and GN method record similar levels of accuracy, and both are clearly better than the Moments method in this respect. 
This is not surprising, as Minka's fixed-point iteration method and the GN method are eventually maximizing the same log-likelihood function.

\pgfplotsset{
	colormap={blueblack}{
		color(0cm)=(black!75!blue);
		color(0.05
		cm)=(blue!75!white);
		color(1cm)=(white)
	}
}
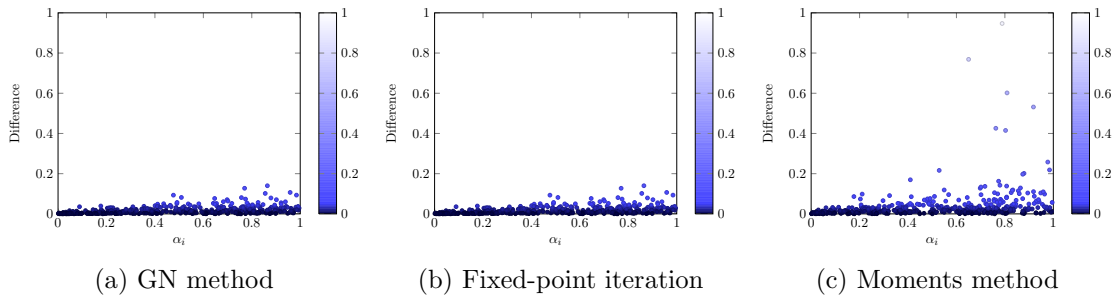
\begin{figure}
	\begin{subfigure}[b]{0.32\textwidth}
		\centering
		\resizebox{\linewidth}{!}{
			\begin{tikzpicture}
				\begin{axis}[ylabel={Difference}, xlabel={$\alpha_i$}, xmin=0,xmax=1, ymin=0, ymax=1, xtick={0,0.2,...,1}, ytick={0,0.2,..., 1}, colorbar, colormap name=blueblack, point meta min=0, point meta max=1]
					\addplot [only marks, scatter, mark=*, mark size=1.5pt] table {Accuracy.Gibbs.10.1.dat};
				\end{axis}
			\end{tikzpicture}
		}
		\caption{GN method}
		\label{fig:accuracy10Dim_1_a}
	\end{subfigure}
	\begin{subfigure}[b]{0.32\textwidth}
	\centering
		\resizebox{\linewidth}{!}{
			\begin{tikzpicture}
				\begin{axis}[ylabel={Difference}, xlabel={$\alpha_i$}, xmin=0,xmax=1, ymin=0, ymax=1, xtick={0,0.2,...,1}, ytick={0,0.2,..., 1}, colorbar, colormap name=blueblack, point meta min=0, point meta max=1]
					\addplot [only marks, scatter, mark=*, mark size=1.5pt] table {Accuracy.Minka.10.1.dat};
				\end{axis}
			\end{tikzpicture}
		}
		\caption{Fixed-point iteration}   
		\label{fig:accuracy10Dim_1_b}
	\end{subfigure}
	\begin{subfigure}[b]{0.32\textwidth}
		\centering
		\resizebox{\linewidth}{!}{
			\begin{tikzpicture}
				\begin{axis}[ylabel={Difference}, xlabel={$\alpha_i$}, xmin=0,xmax=1, ymin=0, ymax=1, xtick={0,0.2,...,1}, ytick={0,0.2,..., 1}, colorbar, colormap name=blueblack, point meta min=0, point meta max=1]
					\addplot [only marks, scatter, mark=*, mark size=1.5pt] table {Accuracy.Ronning.10.1.dat};
				\end{axis}
			\end{tikzpicture}
		}
		\caption{Moments method}
		\label{fig:accuracy10Dim_1_c}
	\end{subfigure}
\caption{The differences between actual and learned values of $\alpha$ parameter components for small values of $\alpha$, $\alpha_i \in \left]0,1\right]$. The smaller the difference the better. (a) The proposed GN method, (b) Minka's fixed-point iteration method, (c) The Moments method} 
\label{fig:accuracy10Dim_1}
\end{figure}

\begin{figure}
	\begin{subfigure}[b]{0.32\textwidth}
		\centering
		\resizebox{\linewidth}{!}{
			\begin{tikzpicture}
				\begin{axis}[ylabel={Difference}, xlabel={$\alpha_i$}, xmin=0,xmax=50, ymin=0, ymax=25, xtick={0,10,...,50}, ytick={0,2,..., 25}, colorbar, colormap name=blueblack, point meta min=0, point meta max=25]
					\addplot [only marks, scatter, mark=*, mark size=1.5pt] table {Accuracy.Gibbs.10.50.dat};
				\end{axis}
			\end{tikzpicture}
		}
		\caption{GN method}
		\label{fig:accuracy10Dim_50_a}
	\end{subfigure}
	\begin{subfigure}[b]{0.32\textwidth}
	\centering
		\resizebox{\linewidth}{!}{
			\begin{tikzpicture}
				\begin{axis}[ylabel={Difference}, xlabel={$\alpha_i$}, xmin=0,xmax=50, ymin=0, ymax=25, xtick={0,10,...,50}, ytick={0,2,..., 25}, colorbar, colormap name=blueblack, point meta min=0, point meta max=25]
					\addplot [only marks, scatter, mark=*, mark size=1.5pt] table {Accuracy.Minka.10.50.dat};
				\end{axis}
			\end{tikzpicture}
		}
		\caption{Fixed-point iteration}   
		\label{fig:accuracy10Dim_50_b}
	\end{subfigure}
	\begin{subfigure}[b]{0.32\textwidth}
		\centering
		\resizebox{\linewidth}{!}{
			\begin{tikzpicture}
				\begin{axis}[ylabel={Difference}, xlabel={$\alpha_i$}, xmin=0,xmax=50, ymin=0, ymax=25, xtick={0,10,...,50}, ytick={0,2,..., 25}, colorbar, colormap name=blueblack, point meta min=0, point meta max=25]
					\addplot [only marks, scatter, mark=*, mark size=1.5pt] table {Accuracy.Ronning.10.50.dat};
				\end{axis}
			\end{tikzpicture}
		}
		\caption{Moments method}
		\label{fig:accuracy10Dim_50_c}
	\end{subfigure}
\caption{The differences between actual and learned values of $\alpha$ parameter components for large values of $\alpha$, $\alpha_i \in \left]0,50\right]$. The smaller the difference the better. (a) The proposed GN method, (b) Minka's fixed-point iteration method, (c) The Moments method} 
\label{fig:accuracy10Dim_50}
\end{figure}
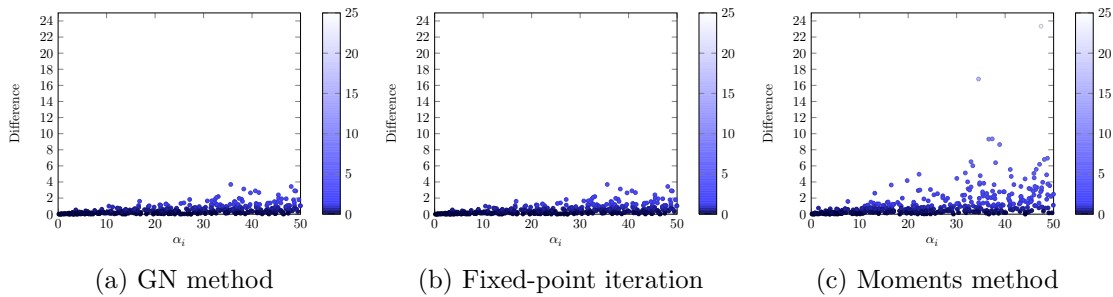

\citet{Wallach08} benchmarks Minka's fixed-point iteration method alongside other methods involving Minka's Newton iteration on the log evidence \citep{Minka00}, fixed-point iteration on the leave-one-out log evidence \citep{Minka00}, and fixed-point iteration on the log evidence introduced by \citet{MacKay94}.
She finds that her efficient implementation of Minka's fixed-point iteration is the fastest and the most accurate.
In this paper, we are comparing the proposed method with Wallach's efficient implementation of Minka's fixed-point iteration method and with the Moments method \citep{Ronning89}.
MALLET \citep{McCallum02} implementation for the Moments method is used.
It can be seen from Figure \ref{fig:accuracy10Dim_1} and Figure \ref{fig:accuracy10Dim_50} that GN and Minka's fixed-point iteration method provide the same level of accuracy. 
However, it is also useful to benchmark those two methods against each other in terms of speed. 

\subsubsection{Speed Benchmark}
Another two data sets are generated for this purpose. 
The first set is generated using a 10 dimensional multivariate Polya distribution whereas the Second set is generated using a 1000 dimensional multivariate Polya distribution.
These two datasets are used to test the performance of the proposed algorithm against Minka's fixed-point iteration in relatively low and high dimensional cases respectively.
Both distributions have known parameter vector $\alpha$, where all components $\alpha_i$ are in $\left]0,1\right]$. 
For both sets, the number of multinomial counts vectors or number of samples falls in the range 10 to 1000, starting from 10 and increasing in steps of 50.
The total number of elements used to generate each sample has a value between 1000 and 20000, starting at 1000 and increasing with step size 1000.  

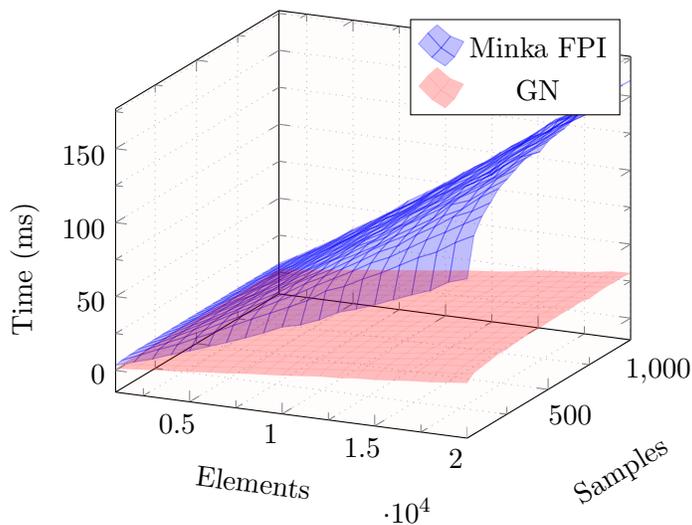
\begin{figure}
	\centering
	\begin{tikzpicture}
    	\begin{axis}[view={25}{21}, grid=both, minor tick num=1, grid style={dotted}, xlabel={Elements}, xlabel style={sloped like x axis}, ylabel={Samples}, ylabel style={sloped}, zlabel = {Time (ms)}, axis background/.style={fill=pink!5!white}] 
			\addplot3[surf, blue, shader=flat, draw opacity=0.30, fill opacity = 0.25] file {Time.Surf.F.C.1.0.10.Minka.dat};
			\addlegendentry{Minka FPI}
			\addplot3[surf, red, shader=flat, draw opacity=0.05, fill opacity = 0.25] file {Time.Surf.F.C.1.0.10.Gibbs.dat};
			\addlegendentry{GN}
		\end{axis}
	\end{tikzpicture}
	\caption{Execution time for GN and Minka's fixed-point iteration (Minka FPI) for a 10 dimensional multivariate Polya distribution using different values of number of samples and different values of number of elements used to generate each sample}
	\label{fig:time10Dim}
\end{figure}

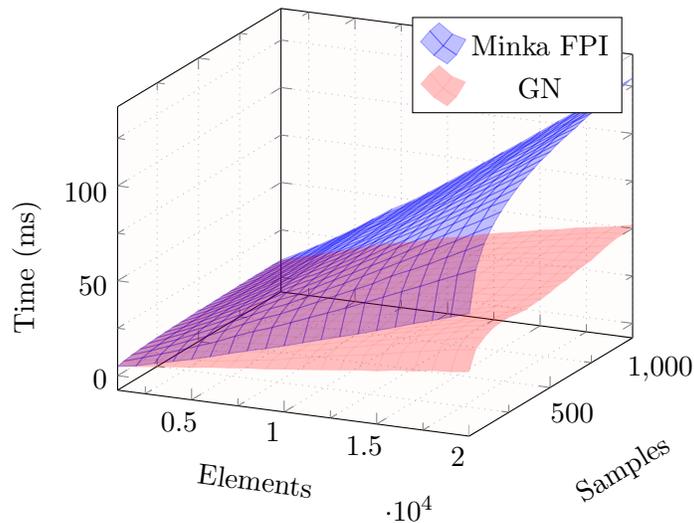
\begin{figure}
	\centering
	\begin{tikzpicture}
    	\begin{axis}[view={25}{21}, grid=both, minor tick num=1, grid style={dotted}, xlabel={Elements}, xlabel style={sloped like x axis}, ylabel={Samples}, ylabel style={sloped}, zlabel = {Time (ms)}, axis background/.style={fill=pink!5!white}]
			\addplot3[surf, blue, shader=flat, draw opacity=0.30, fill opacity = 0.25] file {Time.Surf.F.C.1.0.1000.Minka.dat};
			\addlegendentry{Minka FPI}
			\addplot3[surf, red, shader=flat, draw opacity=0.05, fill opacity = 0.25] file {Time.Surf.F.C.1.0.1000.Gibbs.dat};
			\addlegendentry{GN}
		\end{axis}
	\end{tikzpicture}
	\caption{Execution time for GN and Minka's fixed-point iteration (Minka FPI) for a 1000 dimensional multivariate Polya distribution using different values of number of samples and different values of number of elements used to generate each sample}
	\label{fig:time1000Dim}
\end{figure}

Using the first data set, and for each combination of number of samples and number of elements, the data set is generated from given random $\alpha$ values, and then the time taken by the estimation method is measured.
The solution is considered to be converged (for both methods) when the maximum value among differences between previous guesses of alpha components values and their current estimates is less than 1.0\mbox{\sc{e-6}}.
This process is repeated 100 times, and the mean time is plotted as a dot on the 3D surface shown in Figure \ref{fig:time10Dim}.
The whole process was repeated 100 times again, this time using the higher dimensional samples. The corresponding 3D surface for the high-dimensional trials is shown in Figure \ref{fig:time1000Dim}. 

Figure \ref{fig:time10Dim} and Figure \ref{fig:time1000Dim} show that the proposed method is faster than Minka's fixed-point iteration under all settings. Although the GN method requires more computation inside each iteration over all alpha values, it requires less than half the number of iterations required by Minka's fixed-point iteration method until convergence. 
This speedup is more pronounced in the case of the lower-dimensional dataset, however the number of iterations needed is less than that required by Minka's fixed-point iteration algorithm under all settings.

\section{LDA-GN: Incorporating Hyper-Parameter Inference for Enhanced Topic Models}
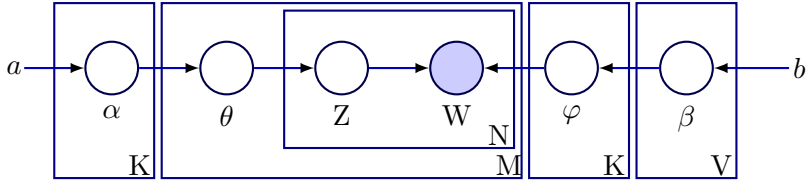
\begin{figure}
	\centering
	\begin{tikzpicture}
		\tikzstyle{variable}=[circle, minimum size = 7mm, thick, draw=blue!30!black, node distance = 8mm,label=below:#1]
		\tikzstyle{hyperparam}=[rectangle, label=center:#1]
		\tikzstyle{connect}=[-latex, thick, draw=blue!75!black]
		\tikzstyle{plate}=[rectangle, draw=blue!50!black, thick, label={[xshift=-14pt,yshift=13pt]south east:#1}]
		\node[hyperparam=$a$] (a) { };
		\node[variable=$\alpha$] (alpha) [right=of a] { };
		\node[variable=$\theta$] (theta) [right=of alpha,] { };
		\node[variable=Z] (Z) [right=of theta] {};
		\node[variable=W, fill = blue!20!white] (W) [right=of Z] { };
		\node[variable=$\varphi$] (phi) [right=of W] { };
		\node[variable=$\beta$] (beta) [right=of phi] { };
		\node[hyperparam=$b$] (b) [right=of beta] { };
		\path (a) edge [connect] (alpha) (alpha) edge [connect] (theta) (theta) edge [connect] (Z) (Z) edge [connect] (W) (phi) edge [connect] (W) (beta) edge [connect] (phi) (b) edge [connect] (beta);
		\node[plate=K, inner xsep=3.0mm, inner ysep=8.0mm, fit = (alpha), xshift=-1mm, yshift=-3mm] {};
		\node[plate=N, inner xsep=4.0mm, inner ysep=5.5mm, fit = (Z) (W), yshift=-1.5mm] {};
		\node[plate=M, inner xsep=5.0mm, inner ysep=8.0mm, fit = (theta) (Z) (W), xshift=0mm, yshift=-3mm] {};
		\node[plate=K, inner xsep=3.0mm, inner ysep=8.0mm, fit = (phi), xshift=1mm, yshift=-3mm] {};
		\node[plate=V, inner xsep=3.0mm, inner ysep=8.0mm, fit = (beta), xshift=0mm, yshift=-3mm] {};
	\end{tikzpicture}
	\caption{LDA-GN model}
	\label{fig:LDA-GN}
\end{figure}

LDA-GN is a variant of LDA that incorporates the proposed GN method, using it to learn variables $\varphi$, $\theta$, $Z$, $\alpha$ and $\beta$. The main idea behind LDA-GN is to allow similar words to have similar beta values and consequently to be distributed similarly over topics. 
Thus, an asymmetric beta prior should be used in this case. 
In order to learn beta values, the LDA model can be extended by placing a non-informative prior before beta variables as shown in Figure \ref{fig:LDA-GN}. 
This gives corpus words the ability to be distributed differently over topics.
This is useful and necessary because some terms need to be participating in a higher number of topics compared with other terms. 
On the other hand, when a symmetric beta is used, all words have to participate in roughly the same number of topics, which can be seen as  a limitation in the original LDA model.
Further, it may be argued that topics should not be bounded by the number of documents that they be distributed over.
Thus, an asymmetric alpha prior is advisable as well.
The same technique is applied to alpha, which is, in other words placing a non-informative prior before the alpha variables, as also shown in Figure \ref{fig:LDA-GN}.

The generative process associated with LDA-GN is described in Algorithm \ref{alg:LDAGNGP}. 
The LDA-GN generative process is similar to the standard LDA generative process, with an extra pair of steps.
The first step is sampling each $\alpha$ vector component value from a uniform distribution with parameters $0$ and $a$.
The second step is sampling each $\beta$ vector component value from a uniform distribution with parameters $0$ and $b$. 
This will give $\alpha_k$ and $\beta_v$ the ability to take any suitable value in the range $[0,a]$ and $[0,b]$ respectively; where $a$ and $b$ are a positive real numbers. 
The remaining steps of the generative process are exactly the same as in the standard LDA model. 

\begin{algorithm}                      
\caption{LDA-GN generative process}
\label{alg:LDAGNGP}
\begin{algorithmic}
    \FOR {$v=1$ \TO $V$}
    	\STATE Choose a beta value $\beta_v \sim Uniform(0,b)$
    \ENDFOR
    \FOR {$k=1$ \TO $K$}
    	\STATE Choose an alpha value $\alpha_k \sim Uniform(0, a)$
    	\STATE Choose a distribution over terms $\varphi_k \sim Dir(\beta)$
    \ENDFOR
    \FOR {$d=1$ \TO $M$}
    	\STATE Draw a topic proportion $\theta_d \sim Dir(\alpha)$
    	\FOR {$t=1$ \TO $N_d$}
    		\STATE Draw a topic assignment $z_{d,n} \sim Multinomial(\theta_d), z_{d,n} \in {1..K}$
    		\STATE Draw a word $w_{d,n} \sim Multinomial(\varphi_{z_{d,n}})$
    	\ENDFOR
    \ENDFOR
\end{algorithmic}
\end{algorithm} 

\subsection{LDA-GN Model Inference}
From Figure \ref{fig:LDA-GN} and the LDA-GN generative process described in Algorithm \ref{alg:LDAGNGP}, the joint distribution is given by the following equation:
\begin{multline}
  P(W,Z,\theta,\varphi,\beta,\alpha|a,b)=\\ \prod_{r=1}^V P(\beta_r|b) \prod_{k=1}^K {P(\alpha_k|a)P(\varphi_k|\beta) \prod_{d=1}^M {P(\theta_d|\alpha) \prod_{t=1}^{N_d} P(Z_{d,t}|\theta_d)P(W_{d,t}|\varphi_{Z_{d,t}})}}\;\;. \nonumber
\end{multline}
Again, the conjugacy between Dirichlet and multinomial distributions allows $\theta$ and $\varphi$ to be marginalized out:
\begin{equation}
  P(W,Z,\beta,\alpha|a,b) = \prod_{k=1}^{K} \frac{1}{a} \prod_{r=1}^{V} \frac{1}{b} \prod_{d=1}^{M}\frac{B(n_{d,\circ}+\alpha)}{B(\alpha)} \prod_{k=1}^{K}\frac{B(n_{\circ}^k+\beta)}{B(\beta)}\;\;. \nonumber
\end{equation}
Gibbs sampling equations:
\begin{equation}
	P(Z_{(d,t)}=k|Z_{\neg(d,t)},W,\alpha,\beta,a,b) \propto (n^{k, \neg (d,t)}_{d,\circ} + \alpha_k)\frac{n^{k,\neg (d,t)}_{\circ,v} + \beta_{v}}{\sum_{r=1}^{V} n^{k,\neg (d,t)}_{\circ,r} + \beta_r} \nonumber
\end{equation}
\begin{equation}
	P\left(\alpha_{k} | \alpha_{\neg k},Z,W,\beta,a,b \right) 
		\propto \prod_{d=1}^{M} \frac{\Gamma(\alpha_{\circ}) \Gamma(n_{d,\circ}^{k}+\alpha_k)}{\Gamma(\alpha_k) \Gamma(n_{d,\circ}^{\circ} + \alpha_{\circ})} 
		\propto \prod_{d=1}^{M} \frac{\prod_{l=0}^{n_{d,\circ}^{k}-1} \alpha_k + l}{\prod_{l=0}^{n_{d,\circ}^{\circ}-1} \alpha_{\circ} + l} \nonumber
\end{equation}
\begin{equation}
	P\left(\beta_{v} | \beta_{\neg v},Z,W,\alpha,a,b \right) 
		\propto \prod_{k=1}^{K} \frac{\Gamma(\beta_{\circ}) \Gamma(n_{\circ,v}^{k}+\beta_v)}{\Gamma(\beta_v) \Gamma(n_{\circ,\circ}^{k} + \beta_{\circ})} 
		\propto \prod_{k=1}^{K} \frac{\prod_{l=0}^{n_{\circ,v}^{k}-1} \beta_v + l}{\prod_{l=0}^{n_{\circ,\circ}^{k}-1} \beta_{\circ} + l}\;\;. \nonumber
\end{equation}
where $n_{\circ,\circ}^k$ is the total number of words assigned to the topic $k$ in the whole corpus, and $n_{d,\circ}^{\circ}$ is the total number of words in the document $W_d$.
Because $P(\theta_d|Z_d, \alpha)$ and $P(\varphi_k|Z, \beta)$ are samples from a Multivariate Polya distribution, Equation \ref{eq:LDA_Gibbs4} and Equation \ref{eq:LDA_Gibbs5} can still be used to calculate the variables $\theta$ and $\varphi$ respectively. 
This calculation can take place after Gibbs sampling convergence by using a good sample. 
Consequently, the LDA-GN collapsed Gibbs sampling algorithm is given by Algorithm \ref{alg:LDAGNCGS}
\begin{algorithm}                      
\caption{LDA-GN collapsed Gibbs sampler}
\label{alg:LDAGNCGS}
\begin{algorithmic}
    \REQUIRE $W$ words of the corpus
    \ENSURE $Z$ topic assignments, $\theta$ topics mixtures, $\varphi$ topics distributions, $\alpha$ and $\beta$ the models parameters.
    \STATE Randomly initialize $Z$ with integers $\in [1..K]$
    \REPEAT
    \FOR {$k = 1$ \TO $K$}
    	\STATE $\alpha_k \leftarrow \argmax{\alpha_k} \left[ \prod_{d=1}^{M} \frac{\Gamma(\alpha_{\circ}) \Gamma(n_{d,\circ}^{k}+\alpha_k)}{\Gamma(\alpha_k) \Gamma(n_{d,\circ}^{\circ} + \alpha_{\circ})}  \right]$
    \ENDFOR
    \FOR {$v = 1$ \TO $V$}
    	\STATE $\beta_v \leftarrow \argmax{\beta_v} \left[ \prod_{k=1}^{K} \frac{\Gamma(\beta_{\circ}) \Gamma(n_{\circ,v}^{k}+\beta_v)}{\Gamma(\beta_v) \Gamma(n_{\circ,\circ}^{k} + \beta_{\circ})} \right]$
    \ENDFOR
    \FOR {$d = 1$ \TO $M$}
		\FOR {$t = 1$ \TO $N_d$} 
			\STATE $v \leftarrow W_{d,t}$; $k \leftarrow Z_{d,t}$
			\STATE $n_{d,\circ}^{k} \leftarrow n_{d,\circ}^{k} - 1$; $n_{\circ,v}^{k} \leftarrow n_{\circ,v}^{k} - 1$; $n_{\circ,\circ}^{k} \leftarrow n_{\circ,\circ}^{k} - 1$; 
			\STATE $k \sim (n^{k}_{d,\circ} + \alpha_k)\frac{n^{k}_{\circ,v} + \beta_{v}}{ n^{k}_{\circ,\circ} + \beta_{\circ}}$
			\STATE $Z_{d,t} \leftarrow k$
			\STATE $n_{d,\circ}^{k} \leftarrow n_{d,\circ}^{k} + 1$; $n_{\circ,v}^{k} \leftarrow n_{\circ,v}^{k} + 1$; $n_{\circ,\circ}^{k} \leftarrow n_{\circ,\circ}^{k} + 1$; 
		\ENDFOR
	\ENDFOR
    \UNTIL {convergence}
    \STATE Calculate $\theta$ using Equation \ref{eq:LDA_Gibbs4}
    \STATE Calculate $\varphi$ using Equation \ref{eq:LDA_Gibbs5}
    \RETURN $Z$,$\theta$,$\varphi$,$\alpha$,$\beta$
\end{algorithmic}
\end{algorithm} 

\section{Evaluation}
Due to the unsupervised nature of LDA-based topic modelling algorithms, evaluation of inferred topic models is a difficult task. 
However, there are some popular methods in the literature to attempt this evaluation. 
A topic model's $perplexity$, under a hold-out set of test documents, is usually used as a standard evaluation metric.  
On the other hand, a topic model's performance in a supervised task can also be used to assess its performance against other models. 
These methods are further described next.

\subsection{Perplexity}
One common way to evaluate a topic model is to calculate its perplexity under a set of unseen test documents. 
Perplexity is a measure to benchmark a topic model's ability to generalize to  unseen documents. 
In other words, it provides a numerical value indicating, in effect, how much the topic model is 'surprised' by new data. 
The higher the probability of test document words given the model, the smaller the perplexity value becomes. 
Consequently, a model with a smaller perplexity value can be considered to have better ability to generalize to unseen documents.
Let $\tilde{W}$ be an unseen test corpus which contains $\tilde{M}$ documents. 
The perplexity is calculated by exponentiating the negative mean log-likelihood value of the whole set of document words given the model.
Perplexity is given by the following formula:

\begin{equation}
	Perplexity(\tilde{W} | W,Z,\alpha,\beta)=\exp\left({\frac{-\sum_{j = 1}^{\tilde{M}} \log P(\tilde{W}_j |  W,Z,\alpha,\beta)}{\sum_{j = 1}^{\tilde{M}} \tilde{N}_j}}\right)
\label{eq:Perplexity}
\end{equation} 
where $j \in {1..\tilde{M}}$ and $\tilde{N}_j$ is the number of words in test document $\tilde{W}_j$. 
Unfortunately the exact value of the marginal distributions $P(\tilde{W}_j | W,Z,\alpha,\beta)$ is intractable due to the need to sum over all different $Z$ settings (Which is exponential in the number of words in the corpus). 
However, there are multiple methods to approximate this marginal probability in the literature such as: annealed importance sampling (AIS) \citep{Neal01}, harmonic mean method \citep{Newton94}, Chib-style estimation \citep{Chib95,Wallach09}.
One of the best methods in the literature is Left-To-Right algorithm \citep{Wallach09,Wray09}.

\subsubsection{The Left-To-Right Algorithm}
The Left-To-Right method is based on breaking the problem into a series of parts:
\begin{multline}
	\begin{split}
	P(\tilde{W}_j|W,Z,\alpha,\beta)
	 		&= \prod_{t=1}^{\tilde{N}_j} P(\tilde{W}_{j,t}|\tilde{W}_{j,1},\tilde{W}_{j,2},...,\tilde{W}_{j, t-1}, W, Z, \alpha, \beta) \\
			&= \prod_{t=1}^{\tilde{N}_j} \sum_{\tilde{Z}_{j,1},...,\tilde{Z}_{j,t}} P(\tilde{W}_{j, t}, \tilde{Z}_{j,1},...,\tilde{Z}_{j,t} |\tilde{W}_{j,1},..., \tilde{W}_{j,t-1}, W,Z, \alpha, \beta)\;\;. \nonumber
	\end{split}
\end{multline}
where $\tilde{Z}_j$ gives the topic assignments of test document $\tilde{W}_j$. 
It can be seen that the previous equation involves marginalizing out the variable $\tilde{Z}_j$;
this is intractable for large test documents and a high number of topics $K$.
Luckily, the previous sum over all possible value settings of $\tilde{Z}_{j,1},...,\tilde{Z}_{j,t}$ can be approximated using sequential Monte Carlo techniques \citep{Moral06} with $R$ particles.
Thus, let $(\tilde{Z}_{j,1},...,\tilde{Z}_{j,t}) \sim P \left(  \tilde{Z}_{j,1},...,\tilde{Z}_{j,t} |\tilde{W}_{j,1},..., \tilde{W}_{j,t-1}, W,Z, \alpha, \beta  \right)$. 
Consequently, the approximation can be calculated using $R$ samples from the previous distribution as follows:
\begin{multline}
	\sum_{\tilde{Z}_{j,1},...,\tilde{Z}_{j,t}} P(\tilde{W}_{j, t}, \tilde{Z}_{j,1},...,\tilde{Z}_{j,t} |\tilde{W}_{j,1},..., \tilde{W}_{j,t-1}, W,Z, \alpha, \beta) \\
		\begin{aligned}	
	 		&\approx \frac{1}{R} \sum_{r = 1}^{R} P(\tilde{W}_{j,t}|\tilde{W}_{j,1},..., \tilde{W}_{j,t-1}, \tilde{Z}_{j,1}^{r},...,\tilde{Z}_{j,t-1}^{r}, W,Z, \alpha, \beta) \\
	 		&= \frac{1}{R} \sum_{r = 1}^{R} \sum_{k = 1}^{K} \frac {n_{\circ, \tilde{W}_{j,t}}^{k} + \beta_{\tilde{W}_{j,t}}}{n_{\circ, \circ}^{k} + \beta_{\circ}} \frac{\tilde{n}_{j,\circ}^{k, r} + \alpha_k}{\sum_{i = 1}^{K} \tilde{n}_{j,\circ}^{i, r} + \alpha_i}\;\;, \nonumber
		\end{aligned}
\end{multline}
where $\tilde{n}_{j,\circ}^{i,r}$ is the number of words in test document $\tilde{W}_j$ that are assigned to topic $i$ in the sample $r$, and $\tilde{Z}_{j,t}^{r}$ is the topic assignment for the $t^{th}$ word in test document $\tilde{W}_j$ and sample $r$.
Thus, the Left-To-Right algorithm is given by Algorithm \ref{alg:LTR}.

\begin{algorithm}                      
\caption{The Left-to-right algorithm to estimate the value $\log P(\tilde{W}_j|W,Z,\alpha,\beta)$}
\label{alg:LTR}
\begin{algorithmic}
    \REQUIRE $W$ words of the training corpus, $\tilde{W}_j$ words of the $j^{th}$ test document, $Z$ topic assignments of the training corpus, $\alpha$ and $\beta$ the models parameters.
    \ENSURE $l = \log P(\tilde{W}_j|W,Z,\alpha,\beta)$ the log likelihood of the test document $\tilde{W}_j$ given a trained LDA model.
    \STATE $l \leftarrow 0$ 
    \FOR {$t = 1$ \TO $\tilde{N}_j$}
    	\STATE $P_t \leftarrow 0$
    	\FOR {$r = 1$ \TO $R$}
    		\FOR {$t^{\prime} = 1$ \TO $t$}
    			\STATE $v \leftarrow \tilde{W}_{j, t^{\prime}}$; $k \leftarrow \tilde{Z}_{j, t^{\prime}}$
    			\STATE $\tilde{n}_{j, \circ}^{k} \leftarrow \tilde{n}_{j, \circ}^{k} - 1$
    			\STATE $k \sim (\tilde{n}_{j, \circ}^{k} + \alpha_k) \frac{n_{\circ, v}^{k} + \beta_v}{n_{\circ, \circ}^{k} + \beta_{\circ}}$
    			\STATE $\tilde{Z}_{j, t^{\prime}} \leftarrow k$
    			\STATE $\tilde{n}_{j, \circ}^{k} \leftarrow \tilde{n}_{j, \circ}^{k} + 1$
    		\ENDFOR
    		\STATE $P_t \leftarrow P_t + \sum_{k = 1}^{K} \frac {n_{\circ, \tilde{W}_{j,t}}^{k} + \beta_{\tilde{W}_{j,t}}}{n_{\circ, \circ}^{k} + \beta_{\circ}} \frac{\tilde{n}_{j,\circ}^{k, r} + \alpha_k}{\sum_{i = 1}^{K} \tilde{n}_{j,\circ}^{i, r} + \alpha_i}$
    	\ENDFOR
    	\STATE $l \leftarrow l + \log \frac{P_t} {R}$
    	\STATE $k \sim (\tilde{n}_{j, \circ}^{k} + \alpha_k) \frac{n_{\circ, \tilde{W}_{j, t}}^{k} + \beta_{\tilde{W}_{j, t}}}{n_{\circ, \circ}^{k} + \beta_{\circ}}$
    	\STATE $\tilde{n}_{j, \circ}^{k} \leftarrow \tilde{n}_{j, \circ}^{k} + 1$
    	\STATE $\tilde{Z}_{j, t} \leftarrow k$
    \ENDFOR
    \RETURN $l$
\end{algorithmic}
\end{algorithm} 

\subsection{Spam Filtering Performance}
Another way to evaluate a topic model is to use the model in a supervised task such as classification or spam filtering. 
The topic model's performance can then be tested against some other models' performance. 
In this paper, we chose a spam filtering task to compare LDA-GN against standard LDA. 
There are multiple ways to use topic model for spam filtering. 
For example, after treating the spam filtering task as a binary classification problem, a topic model can be used as a document dimensionality reduction technique to choose features and then carry out classification using standard methods \citep{Blei03}. 
However, in this paper we use the `Multi-Corpus LDA' approach from \citep{Biro09,Biro08};  that approach is described next.

\subsubsection{Multi-Corpus LDA}
In the Multi-corpus LDA (MC-LDA) approach \citep{Biro09,Biro08}, two distinct LDA models are inferred using the same vocabulary words.  
The First model is inferred from  the collection of spam documents with $K^{(s)}$ topics, whereas the second model is inferred from the collection of a non-spam documents with $K^{(n)}$ topics.
Consequently, the words distributions for $K^{(n)} + K^{(s)}$ topics are learned.
The idea of MC-LDA  is to merge the previous two models and create a unified model with $K^{(n)}+K^{(s)}$ topics.
This is done by simply encoding the topic identification numbers of the spam topic model to begin from $K^{(n)}+1$ instead of beginning from $1$. 
Thus, for an unseen document $\tilde{W}_{\tilde{d}}$, the inference in the unified model can be made using the following formula:
\begin{equation}
  	P(\tilde{Z}_{(\tilde{d},t)}=k|\tilde{Z}_{\neg(\tilde{d},t)},\tilde{W},W,Z,\alpha,\beta) \propto (\tilde{n}^{k, \neg (\tilde{d},t)}_{\tilde{d},\circ} + \alpha_k)\frac{n^{k,\neg (\tilde{d},t)}_{\circ,v} + \beta_{v}}{\sum_{r=1}^{V} n^{k,\neg (\tilde{d},t)}_{\circ,r} + \beta_r}\;\;, \nonumber
\end{equation} 
where $\tilde{n}^{k, \neg (\tilde{d},t)}_{\tilde{d},\circ}$ represents the number of words in test document $\tilde{W}_{\tilde{d}}$ that are assigned to topic $k$ excluding the $t^{th}$ word in that document.
However, the count $n^{k,\neg (\tilde{d},t)}_{\circ,v}$, which represents the number of word instances of vocabulary term $v$ form all documents assigned to topic $k$, is unknown. 
Thus the previous Multi-Corpus inference formula's second factor can be approximated using the $\varphi_{k}^{t}$ value. 
Let $\tilde{W}_{(\tilde{d}, t)}$, which is the $t^{th}$ word in test document $\tilde{W}_{\tilde{d}}$, be $v$, then:
\begin{equation}
  	P(\tilde{Z}_{(\tilde{d},t)}=k|\tilde{Z}_{\neg(\tilde{d},t)},\tilde{W},W,Z,\alpha,\beta) \appropto (\tilde{n}^{k, \neg (\tilde{d},t)}_{\tilde{d},\circ} + \alpha_k)\varphi_{k}^{v}\;\;. \nonumber
\end{equation}
As a result of the inference process and after a sufficient number of iterations, the words topic assignment $Z_{\tilde{d}}$ is calculated.
Consequently, document topic distribution $\theta_{\tilde{d}}$ is calculated using:
\begin{equation}
  	\theta^{k}_{\tilde{d}} = \frac{\tilde{n}^{k}_{\tilde{d},\circ} + \alpha_k}{\sum_{i=1}^{K} \tilde{n}^{i}_{\tilde{d},\circ} + \alpha_i}\;\;. \nonumber
\end{equation}
In order to classify a document $\tilde{W}_{\tilde{d}}$, the LDA prediction value $\tau = \sum_{i=K^{(n)} + 1}^{K^{(s)}} \theta^{i}_{\tilde{d}}$ is calculated.
if the LDA prediction value $\tau$ is above than a specific threshold, the document will be classified as spam.
Otherwise, the document can be classified as legitimate.

\section{Experimental Evaluation of LDA-GN}
In this section, evaluation results using methods detailed in the previous section is displayed. 
This mainly involves evaluating LDA-GN using the perplexity metric and using its performance on a spam filtering task. In both cases, LDA-GN is compared with the standard LDA model suggested in \citep{Wallach09Rethinking}.
According to \citet{Wallach09Rethinking} the standard LDA model use an asymmetric Dirichlet prior over documents-over-topics distributions $\theta$ and a symmetric Dirichlet prior over topics-over-words distributions $\varphi$. 
However, in this paper the new method LDA-GN has asymmetric Dirichlet priors over both documents-over-topics distributions $\theta$ and topics-over-words distributions $\varphi$. 

\subsection{Perplexity Score}
Using Algorithm \ref{alg:LDAGNCGS}, an LDA-GN model Gibbs sampler is implemented.  
On the other hand, MALLET \citep{McCallum02} LDA implementation is used for standard LDA. 
Both  are implemented using Java. Recommended settings suggested by \citet{Wallach09Rethinking} are used for the standard LDA model, which are: asymmetric Dirichlet prior over documents-over-topics distributions and a symmetric Dirichlet prior over topics-over-words distributions.

In order to train and evaluate these models, two corpora are used.
First corpus is EPSRC corpus (623 documents containing 122672 words and 13035 vocabularies) \citep{Khalifa13a} which comprises summaries of projects in Information and Communication
Technology (ICT) funded by the Engineering and Physical Sciences Research Council (EPSRC).
Second corpus is News corpus (2213 documents containing 453462 words and 38500 vocabularies) which is a subset of Associated Press (AP) data from the First Text Retrieval Conference (TREC-1) \citep{Harman93a}. 
Both corpora are provided at \href{http://is.gd/GNTMOD}{\url{http://is.gd/GNTMOD}}.
All standard English stop words are removed from the corpora before learning or inference application.
Each corpus is divided into two parts: the first part is used for training, whereas the second part is used for evaluation purposes.
The first part, which comprises 80\% of corpus documents, is used to train both LDA and LDA-GN models. The remaining 20\% part are used to calculate perplexity scores using Equation \ref{eq:Perplexity}. 
In order to calculate probabilities $P(\tilde{W}_j|W,Z,\alpha,\beta)$, a Java implementation of the Left-To-Right algorithm \ref{alg:LTR} is used,
where $j \in [1..\tilde{M}]$ is the test document index, $\tilde{W}_j$ is the $j^{th}$ test document and $\tilde{M}$ is total number of test documents.
Thus, a better model should has a higher probability $P(\tilde{W}_j|W,Z,\alpha,\beta)$ value and consequently a lower perplexity score.

Initial values of variable $\alpha$ are set as $\alpha_k=50/K$ for all topics $k \in [1..K]$.  The $\beta$ variable values are initialized as $\beta_v = 0.01$ for all vocabulary terms $v \in [1..V]$. 
These initial values are recommended in the MALLET package documentation \citep{McCallum02}.
After that, the standard LDA model's MALLET implementation is run using a training corpus as an input.
For the first 50 iterations (the burn-in period), both $\alpha$ and $\beta$ values are kept fixed.
After the burn-in period, Minka's fixed-point iteration is used to learn $\alpha$ and $\beta$ values from the sampler's histograms. 
The $\alpha$ and $\beta$ values learning process is repeated once every 20 iterations. 
After 2000 iterations, the model is considered fully trained.
On the other hand, LDA-GN model is trained using the same training corpus which is used for standard LDA. 
Asymmetric $\alpha$ and $\beta$ values are used in this model. 
Similarly to the standard LDA model, LDA-GN model is considered fully trained after 2000 iterations.

\begin{figure}
\centering
\includegraphics[height=1.82in, width=3.36in]{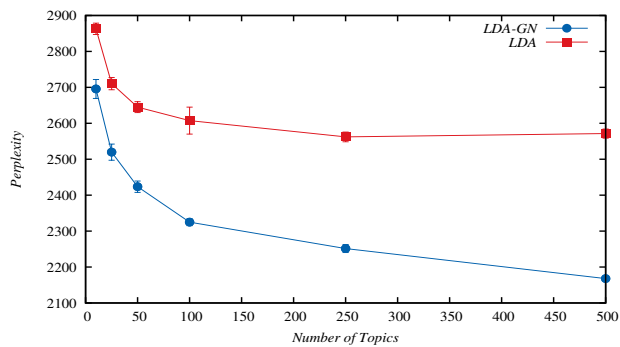} 
\caption{EPSRC corpus, LDA and LDA-GN perplexity values for different number of topics}
\label{fig:EPSRC_Perp}
\end{figure}

\begin{figure}
\centering
\includegraphics[height=1.82in, width=3.36in]{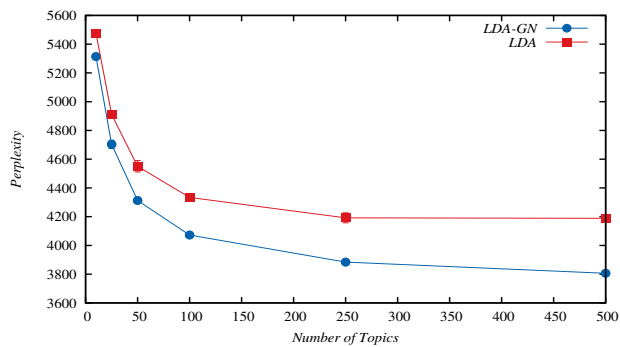} 
\caption{NEWS corpus, LDA and LDA-GN perplexity values for different number of topics}
\label{fig:NEWS_Perp}
\end{figure}

The performance of standard LDA and LDA-GN is tested over a range of scenarios. Both approaches are run five times for each of the following settings for number of topics: 5, 10, 25, 50, 100, 150, 200, 300, and 600 topics. Following every individual run,a fresh split is used to generate training and testing corpora.
Figure \ref{fig:EPSRC_Perp} and Figure \ref{fig:NEWS_Perp} show perplexity values of unseen test data for models inferred by LDA-GN and LDA, on the EPSRC and News corpora respectively. 
Error bars are drawn for each point in the figures. 
Figure \ref{fig:EPSRC_Perp} and Figure \ref{fig:NEWS_Perp} show that LDA-GN outperforms standard LDA for all settings in these two corpora used for evaluation, suggesting that topic models inferred via LDA-GN are better able to generalize than model's inferred via standard LDA.

\subsection{Spam Filtering Score}
Another way to evaluate a topic model is to check its performance in a supervised task such as spam filtering.
Thus, two spam filters are built using the MC-LDA method elaborated before.
The first one  is built using standard LDA whereas the second one is built using LDA-GN. 
Three spam corpora are used for evaluation purposes: (i) The Enron corpus \citep{Metsis06}, which comprises a subset of Enron emails from the period from 1999 until 2002; this corpus contains 16545 legitimate message and 17169 spam;(ii) the LingSpam corpus \citep{Sakkis03} which contains 2412 legitimate message and 481 spam; (iii) The SMS Collection v.1 \citep{Almeida11} which contains 4827 legitimate SMS messages and 747 spam SMS messages.
Standard English stop words are removed from these three corpora.
Each corpus is split into two parts: the first part, which comprises 80\% of the corpus, is used for training whereas the remaining 20\% is used for testing purposes.
using only the training part, two MC-LDA models are built using standard LDA and LDA-GN respectively. 

The first MC-LDA model which is built using standard LDA comprises two LDA models combined. 
The first one is trained using only legitimate messages and 50 topics, whereas the second one is trained using only spam messages and 10 topics.
On the other hand, a second MC-LDA model which is built using LDA-GN comprises two LDA-GN models combined. 
Again, the first one is trained using legitimate messages and 50 topics, whereas the second one is trained using spam messages with 10 topics.
Given two fully trained MC-LDA models, an inference is performed for all test documents.
In order to fully test the models' classification abilities, multiple thresholds are used. 
Thresholds values used are: 0.05, 0.1, 0.25, 0.3, 0.35, 0.4, 0.5, 0.6, 0.7, 0.8 and 0.9.

\begin{figure}[t!]
	\begin{subfigure}[t]{0.5\textwidth}
		\centering
		\includegraphics[height=1.5in]{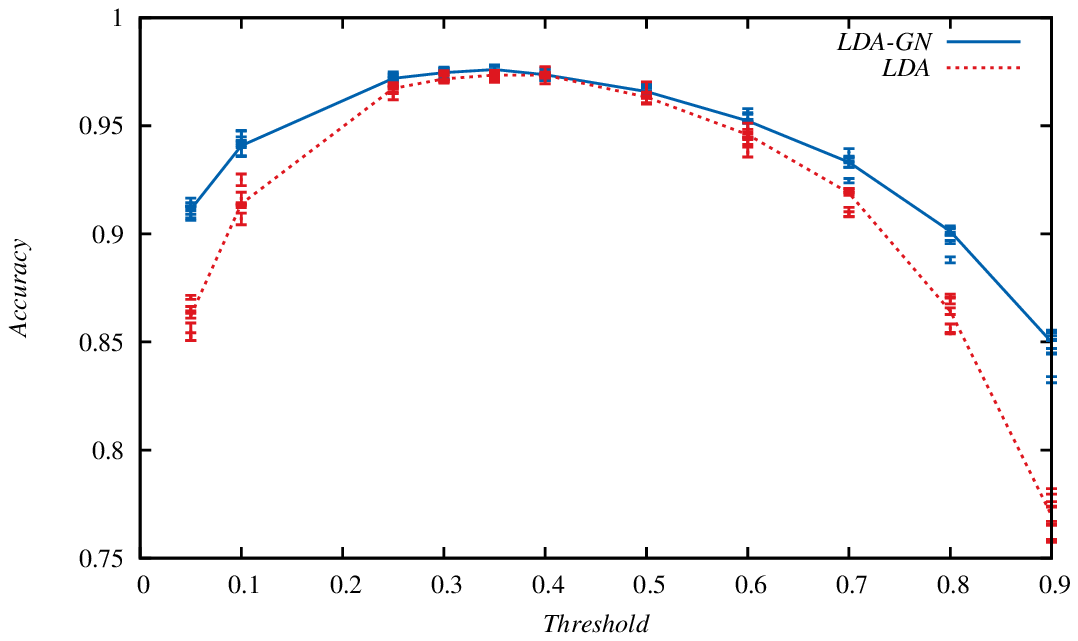}{a}
		\caption{Spam filtering accuracy}
		\label{fig:Enron_Acc}
	\end{subfigure}
	~
	\begin{subfigure}[t]{0.5\textwidth}
		\centering
		\includegraphics[height=1.5in]{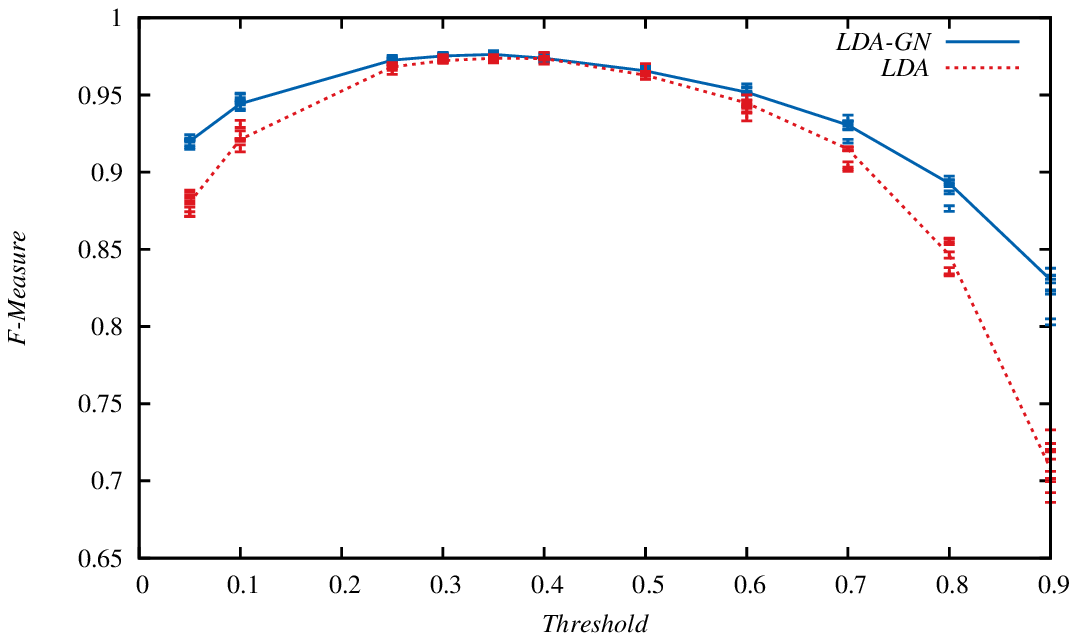}{b}
		\caption{Spam filtering f-Measure}   
		\label{fig:Enron_fmsr}
	\end{subfigure}
	\caption{Enron corpus, LDA and LDA-GN spam filtering performance using different threshold settings} 
	\label{fig:Enron_Performance}
\end{figure}

\begin{figure}[t!]
	\begin{subfigure}[t]{0.5\textwidth}
		\centering
		\includegraphics[height=1.5in]{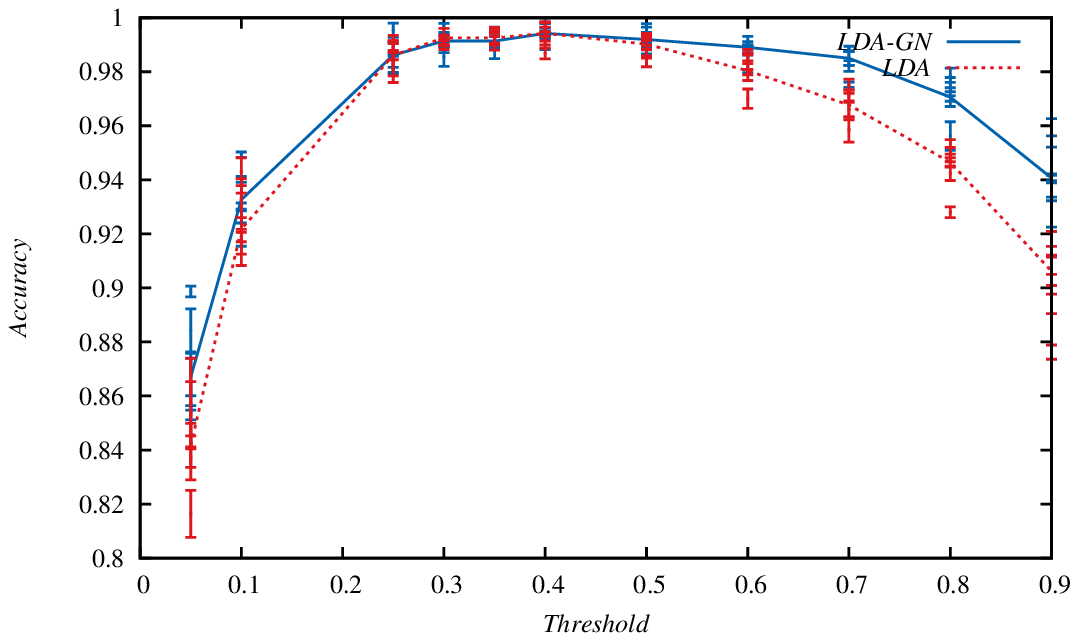}{a}
		\caption{Spam filtering accuracy}
		\label{fig:LingSpam_Acc}
	\end{subfigure}
	~
	\begin{subfigure}[t]{0.5\textwidth}
		\centering
		\includegraphics[height=1.5in]{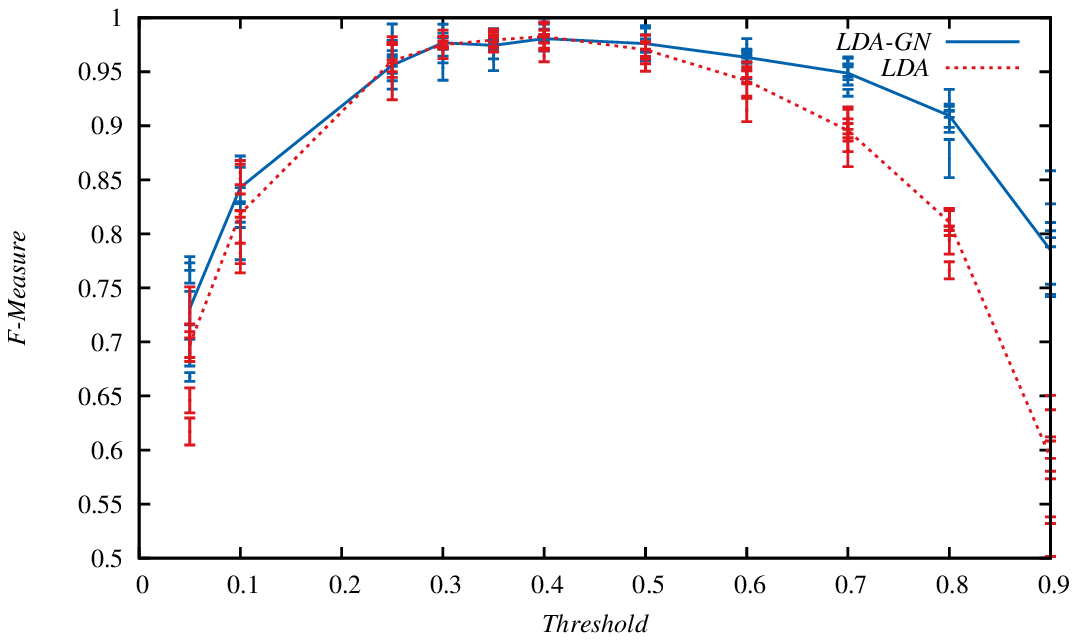}{b}
		\caption{Spam filtering f-Measure}   
		\label{fig:LingSpam_fmsr}
	\end{subfigure}
	\caption{LingSpam corpus, LDA and LDA-GN spam filtering performance using different threshold settings} 
	\label{fig:LingSpam_Performance}
\end{figure}

\begin{figure}[t!]
	\begin{subfigure}[t]{0.5\textwidth}
		\centering
		\includegraphics[height=1.5in]{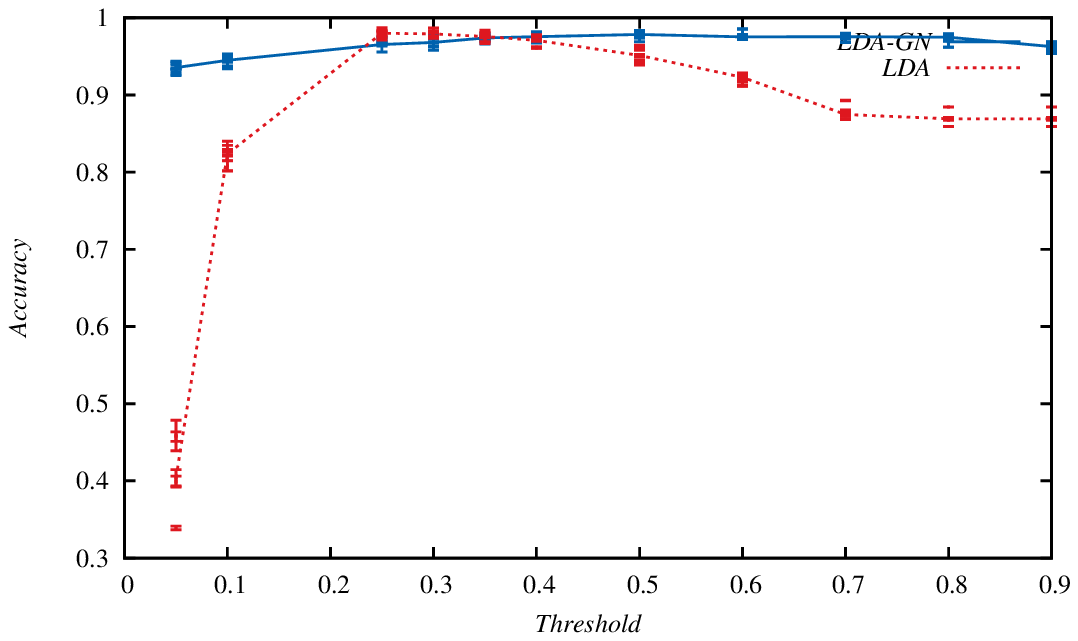}{a}
		\caption{Spam filtering accuracy}
		\label{fig:SMSSpam_Acc}
	\end{subfigure}
	~
	\begin{subfigure}[t]{0.5\textwidth}
		\centering
		\includegraphics[height=1.5in]{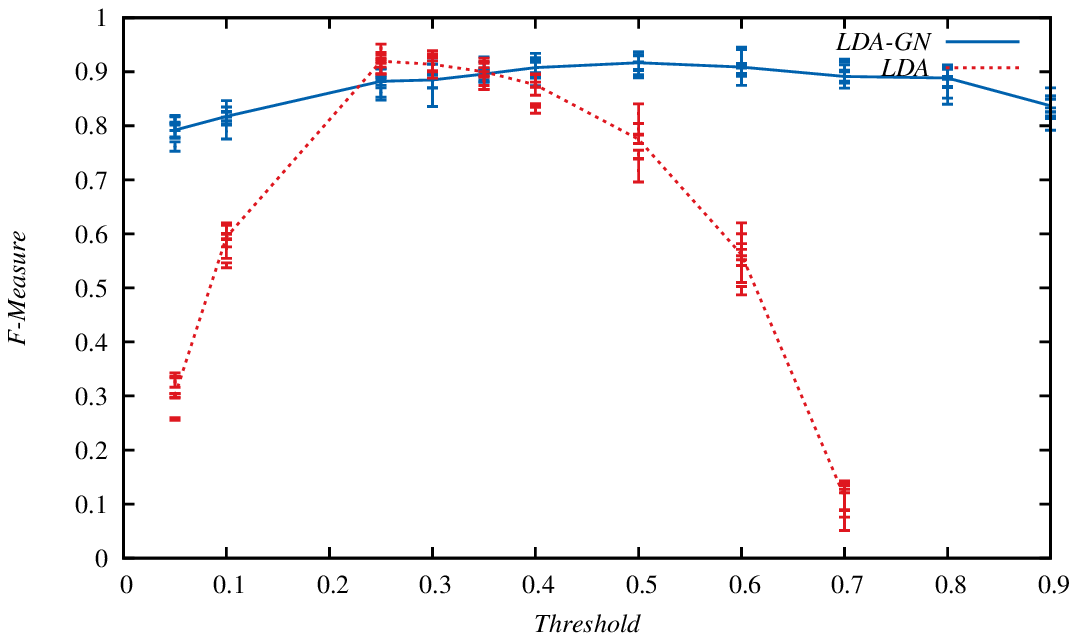}{b}
		\caption{Spam filtering f-Measure}   
		\label{fig:SMSSpam_fmsr}
	\end{subfigure}
	\caption{SMS Collection v.1 corpus, LDA and LDA-GN spam filtering performance using different threshold settings} 
	\label{fig:SMSSpam_Performance}
\end{figure}

For each threshold and given the trained MC-LDA models the inference is applied three times for each model.
Mean values of accuracy and f-Measure are calculated, then these points are registered in a graph.
Standard deviation or standard error (SEM) values of accuracy and f-Measure are calculated as well, and shown as error bars.
The whole process is repeated 5 times, every time with a fresh train/test split.
Eventually, the median of the  five points associated with each threshold value is calculated and a curve is drawn.
Figure \ref{fig:Enron_Acc}, Figure \ref{fig:LingSpam_Acc} and Figure \ref{fig:SMSSpam_Acc} show accuracy scores for both LDA-GN and standard LDA models for the Enron, LingSpam and SMS Collection v.1 corpora respectively.
Moreover, Figure \ref{fig:Enron_fmsr}, Figure \ref{fig:LingSpam_fmsr} and Figure \ref{fig:SMSSpam_fmsr} show f-Measure scores for both LDA-GN and standard LDA models for the Enron, LingSpam and SMS Collection v.1 corpora respectively.
Perusal of these figures shows that models inferred via LDA-GN lead to results that are less sensitive to the threshold value.
However, when the right threshold value is chosen both, models are able to provide almost the same level of accuracy.

\begin{figure}[t!]
	\begin{subfigure}[t]{0.5\textwidth}
		\centering
		\includegraphics[height=1.5in]{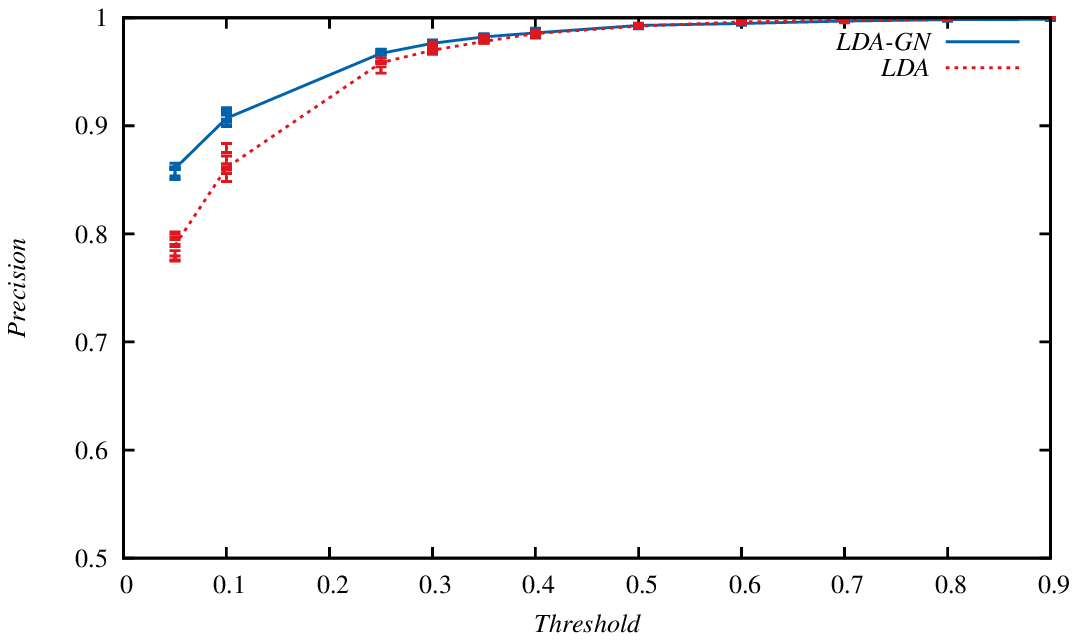}{a}
		\caption{Spam filtering precision}
		\label{fig:Enron_prcsn}
	\end{subfigure}
	~
	\begin{subfigure}[t]{0.5\textwidth}
		\centering
		\includegraphics[height=1.5in]{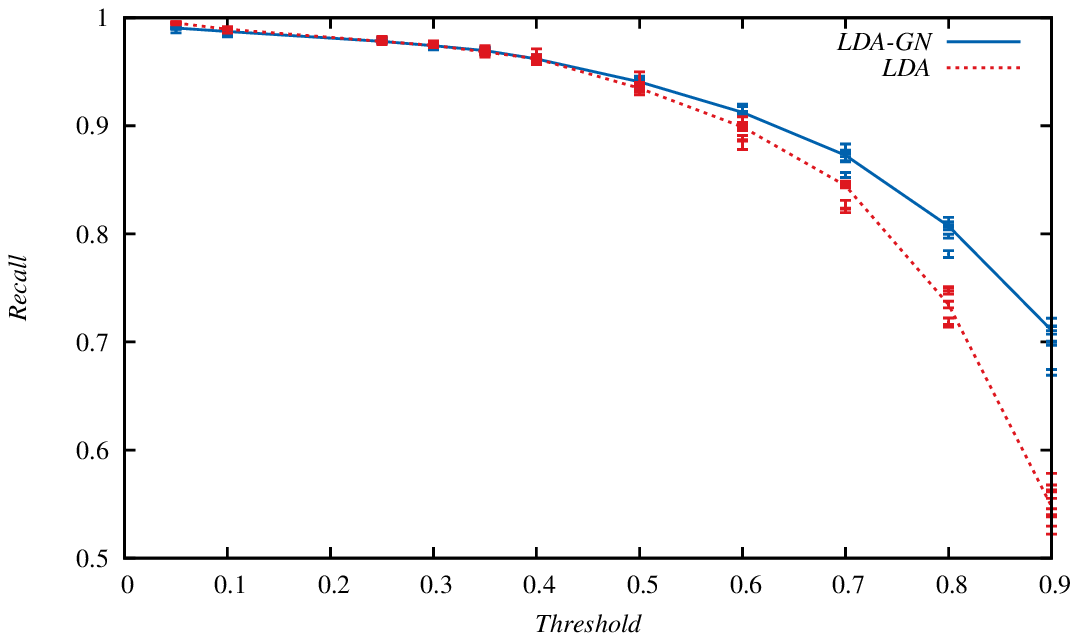}{b}
		\caption{Spam filtering recall}   
		\label{fig:Enron_recal}
	\end{subfigure}
	\caption{Enron corpus, LDA and LDA-GN spam filtering precision and recall scores using different threshold settings} 
	\label{fig:Enron_Performance2}
\end{figure}

Since LDA-GN models provide less sensitivity to threshold values, it can be argued that the topic models inferred by LDA-GN have higher discrimination than those inferred by standard LDA. 
In the MC-LDA approach, a document is classified as spam if its score $\tau = \sum_{i=K^{(n)} + 1}^{K^{(s)}} \theta^{i}_{\tilde{d}}$ is larger than the specific threshold value.  
On the one hand, when the threshold value is less than its optimal value, the spam filter tends to become more strict.
This means that more legitimate documents are classified as spam. 
On the other hand, when the threshold value is larger than its optimal value, the spam filter tends to become more tolerant, consequently, classifying more spam documents  as legitimate.
MC-LDA based on LDA-GN is able to achieve higher accuracy and f-measure scores than MC-LDA based on LDA model when the threshold is less or more than its optimal value. 
More specifically, Figure \ref{fig:Enron_Performance2} shows that LDA-GN, in the Enron case, has better precision for threshold values less than optimal, and better recall scores for threshold values more than optimal. 
Thus, a legitimate document's score over LDA-GN spam topics is always less than its score over standard LDA spam topics. 
On the contrary, a spam document's score over LDA-GN spam topics is always higher than its score over standard LDA spam topics.
So, the topic model inferred by LDA-GN provides a better representation of the corpus than that inferred by standard LDA.

\section{Conclusions}
In this paper, two main contributions are offered: Firstly, a new algorithm to learn multivariate Polya distribution parameters named `GN' is described and evaluated.
Secondly, based on GN, a new extension for LDA, dubbed `LDA-GN' is proposed and evaluated.
  
In order to assess its performance, GN is compared with two other appropriate methods: The Moments method---a quick and approximate approach---and Minka's fixed-point iteration method---a more accurate and a slower method. 
GN is able to infer more accurate values than the Moments method and it is able to provide the same level of accuracy provided by the Minka's fixed-point iteration method.
However, the time taken by GN to compute its results is invariably less than the time consumed by the Minka's fixed-point iteration method for the same accuracy.
GN algorithm can be used in all applications that use the Dirichlet distribution or multivariate Polya distributions to learn parameters from the data itself.

The new extension LDA-GN shows better performance compared with standard LDA. Our experiments using two corpora suggest that its ability to generalise to unseen documents is greater, since it shows lower perplexity values over unseen documents.  
To measure how both models perform in a supervised task, standard LDA and LDA-GN were used in the context of the MC-LDA method in a spam classification task.  
Generally, LDA-GN showed  better performance in this task over multiple choices of the threshold value. 
However, both models were are able to provide the same levels of accuracy given judicious choices of the threshold value.

The lower sensitivity to threshold in the spam classification tasks---as shown by models inferred using LDA-GN---suggests that LDA-GN was able to infer higher quality topic models than LDA, being better representations and more discriminatory of the legitimate and spam parts of these corpora. 
 
Recommended settings described in \citep{Wallach09Rethinking}, which are mainly using asymmetric alpha and symmetric beta priors, lead to different words generally being constrained to contribute to the same number of topics. 
When a symmetric $beta$ is used, and all $beta_v$ have a relatively large value, some words that should really only appear in a small number of topics are encouraged to spread to other topics. 
On the other hand, when $beta_v$ has a relatively small value, all words tend to be distributed over a small number of topics, despite the fact that some words could legitimately appear in many more topics. 
Consequently, topic models built with these constraints can typically contain many irrelevant words among the topics.
In contrast, in LDA-GN every vocabulary term has the freedom to be distributed over any number of topics with no restriction.
However, with no such restriction, stop words will be encouraged to be distributed over all topics evenly. 
So, it is important to remove stop words before an LDA-GN model is learnt. 
That is why all stop words were removed in advance for both LDA and LDA-GN in this paper.

One potential area of future work for LDA-GN is to investigate the placement of  informed priors before the alpha and beta variables. 
Such may be available for many applications (including, for example, updating a topic model following extension of the corpus).  
Meanwhile, the quality of the topic models learned by LDA-GN seems to augur well for their use in supervised learning tasks; spam classification is one example, but we believe other tasks in the general area of supervised document classification may benefit from LDA-GN in the context of the MC-LDA approach. 
This may be especially fruitful in the case of discrimination tasks that involve 'close' categories (e.g. 'finance' vs 'insurance').

\newpage

\bibliography{khalifa15a}

\end{document}